\documentclass[10pt,twocolumn,letterpaper]{article}
\usepackage[pagenumbers]{iccv}      

\definecolor{iccvblue}{rgb}{0.21,0.49,0.74}
\usepackage[pagebackref,breaklinks,colorlinks,allcolors=iccvblue]{hyperref}

\usepackage[capitalize]{cleveref}
\crefname{app}{appendix}{appendices}
\Crefname{app}{Appendix}{Appendices}

\usepackage{tikz}
\usepackage[outline]{contour}
\contourlength{1.5pt}

\usepackage{wrapfig}
\usepackage{subcaption}

\usepackage{bbm}
\usepackage{physics}

\usepackage{mathtools, nccmath}

\usepackage{makecell}

\usetikzlibrary{decorations.pathreplacing}
\usetikzlibrary{calligraphy}
\usetikzlibrary{calc}
\usetikzlibrary{arrows.meta,arrows}
\usetikzlibrary{shapes.geometric}

\newcommand{\toappendix}[1]{}

\def\st{\emph{s.t}\onedot}

\DeclareMathOperator{\R}{\mathbb{R}}

\DeclareMathOperator{\normal}{\mathcal{N}}
\DeclareMathOperator{\noisedist}{\mathcal{Q}}
\DeclareMathOperator{\prob}{\mathbb{P}}
\DeclareMathOperator{\cond}{\,|\,}
\DeclareMathOperator{\likeli}{\mathcal{L}}
\DeclareMathOperator{\data}{\mathcal{D}}

\DeclareMathOperator{\img}{\mathcal{I}}
\DeclareMathOperator{\cell}{\mathnormal{x}}
\DeclareMathOperator{\cells}{\mathcal{X}}

\DeclareMathOperator{\assignment}{\mathnormal{A}}
\DeclareMathOperator{\assignments}{\mathcal{A}*}
\DeclareMathOperator{\optassignment}{\mathnormal{A_{\text{opt}}}}

\DeclareMathOperator*{\argmax}{\arg\max}

\DeclareMathOperator{\nn}{\mathnormal{f_{\theta}}}

\DeclareMathOperator{\tempparam}{{\color{magenta} \tau}}

\newcommand{\ind}[1]{\mathbbm{1}_{\{#1\}}}
\newcommand{\gurobi}{\texttt{Gurobi}}

\usetikzlibrary{patterns, arrows}
\tikzset{
    halo/.style={
        preaction={
            draw,
            white,
            line width=4pt,
            shorten >=1\pgflinewidth,
            cap=round,
            -
        },
        preaction={
            draw,
            white,
            line width=2pt,
            shorten >=-0.1\pgflinewidth
        },
        preaction={
            draw,
            white,
            line width=2pt,
            shorten >=-0.3\pgflinewidth
        },
        preaction={
            draw,
            white,
            line width=2pt,
            shorten >=-0.5\pgflinewidth
        },
        preaction={
            draw,
            white,
            line width=2pt,
            shorten >=-0.7\pgflinewidth
        },
        preaction={
            draw,
            white,
            line width=2pt,
            shorten >=-0.9\pgflinewidth
        }
    }
}

\definecolor{tabblu}{HTML}{1f77b4}
\definecolor{taborg}{HTML}{ff7f0e}
\definecolor{tabgre}{HTML}{2ca02c}
\definecolor{tabred}{HTML}{d62728}
\definecolor{tabppl}{HTML}{9467bd}
\definecolor{tabbrw}{HTML}{8c564b}
\definecolor{tabpnk}{HTML}{e377c2}
\definecolor{tabgry}{HTML}{7f7f7f}
\definecolor{tabolv}{HTML}{bcbd22}
\definecolor{tabcya}{HTML}{17becf}

\definecolor{vanilla}{HTML}{008066}
\definecolor{TS}     {HTML}{ff0000}
\definecolor{FP}     {HTML}{78bc66}
\definecolor{FPTS}   {HTML}{ff7800}
\definecolor{FPA}    {HTML}{aad466}
\definecolor{FPATS}  {HTML}{ffaa00}
\definecolor{SA}     {HTML}{d1e866}
\definecolor{SATS}   {HTML}{ffd100}

\def\width{6pt}
\def\height{3pt}

\newcommand*{\SMm}{%
    \raisebox{.3ex}{%
        \begin{tikzpicture}%
            \node[%
                minimum width=\width,%
                minimum height=\height,%
                fill=vanilla,%
                inner sep=1pt] {};%
        \end{tikzpicture}%
    }%
    \xspace%
}%
\newcommand*{\SMTSm}{%
    \raisebox{.3ex}{%
        \begin{tikzpicture}%
            \node[%
                minimum width=\width,%
                minimum height=\height,%
                fill=TS,%
                inner sep=1pt] {};%
        \end{tikzpicture}%
    }%
    \xspace%
}%
\newcommand*{\FPm}{%
    \raisebox{.3ex}{%
        \begin{tikzpicture}%
            \node[%
                minimum width=\width,%
                minimum height=\height,%
                fill=FP,%
                inner sep=1pt] {};%
        \end{tikzpicture}%
    }%
    \xspace%
}%
\newcommand*{\FPTSm}{%
    \raisebox{.3ex}{%
        \begin{tikzpicture}%
            \node[%
                minimum width=\width,%
                minimum height=\height,%
                fill=FPTS,%
                inner sep=1pt] {};%
        \end{tikzpicture}%
    }%
    \xspace%
}%
\newcommand*{\FPAm}{%
    \raisebox{.3ex}{%
        \begin{tikzpicture}%
            \node[%
                minimum width=\width,%
                minimum height=\height,%
                fill=FPA,%
                inner sep=1pt] {};%
        \end{tikzpicture}%
    }%
    \xspace%
}%
\newcommand*{\FPATSm}{%
    \raisebox{.3ex}{%
        \begin{tikzpicture}%
            \node[%
                minimum width=\width,%
                minimum height=\height,%
                fill=FPATS,%
                inner sep=1pt] {};%
        \end{tikzpicture}%
    }%
    \xspace%
}%
\newcommand*{\ASm}{%
    \raisebox{.3ex}{%
        \begin{tikzpicture}%
            \node[%
                minimum width=\width,%
                minimum height=\height,%
                fill=SA,%
                inner sep=1pt] {};%
        \end{tikzpicture}%
    }%
    \xspace%
}%
\newcommand*{\ASTSm}{%
    \raisebox{.3ex}{%
        \begin{tikzpicture}%
            \node[%
                minimum width=\width,%
                minimum height=\height,%
                fill=SATS,%
                inner sep=1pt] {};%
        \end{tikzpicture}%
    }%
    \xspace%
}%
\newcommand{\SM}{SM~\SMm}%
\newcommand{\SMTS}{SM+TS~\SMTSm}%
\newcommand{\FP}{FP~\FPm}%
\newcommand{\FPTS}{FP+TS~\FPTSm}%
\newcommand{\FPA}{FP+A~\FPAm}%
\newcommand{\FPATS}{FP+A+TS~\FPATSm}%
\newcommand{\AS}{AS~\ASm}%
\newcommand{\ASTS}{AS+TS~\ASTSm}%
\newcommand*{\temp}{%
    \raisebox{.3ex}{%
        \begin{tikzpicture}[x=1.7pt]%
            \node[%
                minimum width=.25\width,%
                minimum height=\height,%
                fill=TS,%
                inner sep=1pt] at (0, 0) {};%
            \node[%
                minimum width=.25\width,%
                minimum height=\height,%
                fill=FPTS,%
                inner sep=1pt] at (1, 0) {};%
            \node[%
                minimum width=.25\width,%
                minimum height=\height,%
                fill=FPATS,%
                inner sep=1pt] at (2, 0) {};%
            \node[%
                minimum width=.25\width,%
                minimum height=\height,%
                fill=SATS,%
                inner sep=1pt] at (3, 0) {};%
        \end{tikzpicture}%
    }%
    \xspace%
}%

\title{How To Make Your Cell Tracker Say ``I dunno!''}

\author{
    Richard D. Paul$^{1,2}$ \quad 
    Johannes Seiffarth$^{1,3}$ \quad 
    David R\"ugamer$^{2,4}$ \quad
    Katharina N\"oh$^1$ \quad 
    Hanno Scharr$^1$ 
    \\[.5mm]
    $^1$ Forschungszentrum J\"ulich \quad
    $^2$ LMU Munich \quad
    $^3$ RWTH Aachen University \\
    $^4$ Munich Center for Machine Learning
    \\
    {\tt\small \{r.paul,j.seiffarth,h.scharr,k.noeh\}@fz-juelich.de} \quad 
    {\tt\small david.ruegamer@stat.uni-muenchen.de}
}

\begin{document}
\maketitle

\begin{abstract}
    Cell tracking is a key computational task in live-cell microscopy,
    but fully automated analysis of high-throughput imaging requires
    reliable and, thus, uncertainty-aware data analysis tools, as the amount of 
    data recorded within a single experiment
    exceeds what humans are 
    able to overlook.
    We here propose and benchmark various methods to reason about and 
    quantify uncertainty in linear assignment-based cell tracking algorithms.
    Our methods take inspiration from statistics and machine learning, 
    leveraging two perspectives on the cell tracking problem 
    explored throughout this work: 
    Considering it as a Bayesian inference problem and as a classification 
    problem.
    Our methods admit a framework-like character in that they equip
    any frame-to-frame tracking method with
    uncertainty quantification.
    We demonstrate this by applying it to various existing tracking algorithms 
    including the recently presented Transformer-based trackers.
    We demonstrate empirically that our methods yield useful and 
    well-calibrated tracking uncertainties.
\end{abstract}

\section{Introduction}
\label{sec:introduction}

Uncertainty-aware cell tracking is a key requirement for fully automated
data analysis of high throughput live-cell microscopy (LCM) data, 
where images often contain
hundreds or thousands of almost indistinguishably looking, moving, growing and dividing cells and where
temporal resolution of the time-lapses is limited by biological and technical considerations
such as phototoxicity \cite{tinevez_chapter_2012} or 
camera movement speed in multi-colony setups \cite{seiffarth_tracking_2024}.
LSCM enables researchers to analyze cellular behavior
beyond the population level, revealing dynamics, development and multi-species 
interactions \cite{blobaum_quantifying_2024, fante_time-resolved_2024, burmeister_optochemical_2021}.
The recent advances in computer vision, mainly driven by the success of deep 
learning and ever-improving computational resources, form a substantial pillar
of modern LCM analyses, as the amount of data collected 
in a single 
experiment 
exceeds what humans are able to overlook 
\cite{kasahara_enabling_2023, seiffarth_tracking_2024, witting_microfluidic_2025}.

Commonly, a data analysis pipeline for LCM data consists of a two-step procedure 
referred to as the \emph{Tracking-by-Detection} (TbD) paradigm 
\cite{scherr_cell_2020,loffler_graph-based_2021,ruzaeva_cell_2022,gallusser_trackastra_2024, fukai_laptrack_2023}.
Cells first get
detected and segmented using a deep learning model 
\cite{ronneberger_u-net_2015, scherr_cell_2020, upschulte_contour_2022, stringer_cellpose_2021, cutler_omnipose_2022,lee_mediar_2022} 
before being tracked
across a sequence of images.
Up until recently,  
tracking 
was most often solved using hand-crafted
heuristics like nearest neighbor associations or linear assignments 
on overlap- or distance-based cost matrices.
More recent methods 
replace such hand-crafted cost functions by ones learned from
data \cite{ben-haim_graph_2022, gallusser_trackastra_2024}, 
having shown great
generalization performance.

\begin{figure}
    \centering
    \resizebox{\columnwidth}{!}{
    \begin{tikzpicture}[x=1in,y=1in]
    \node at (-0.7, 0) {$A_2$:};
    
    \node[transform shape] at(2.2, 0) {\includegraphics[width=1in]{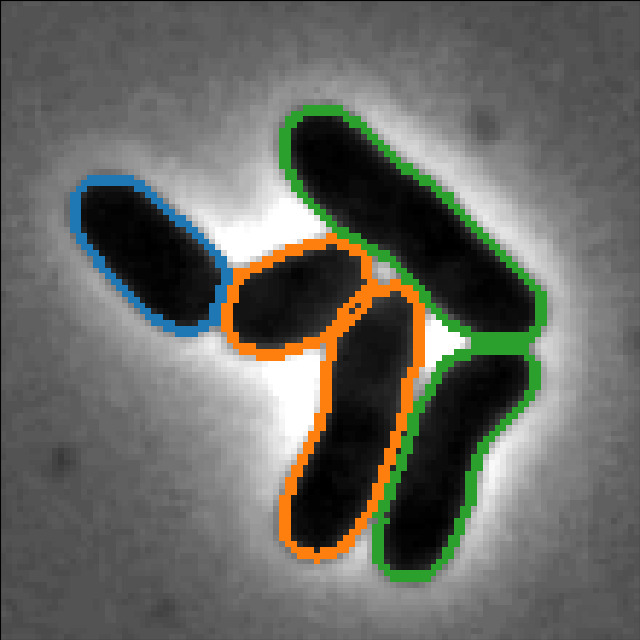}};

    \node[circle,minimum size=5px,inner sep=0,outer sep=0,fill=taborg] (d1) at  (0.55-.5+2.2,-0.66+.5) {};
    \node[circle,minimum size=5px,inner sep=0,outer sep=6pt,fill=taborg] (d2) at  (0.45-.5+2.2,-0.46+.5) {};
    \node[circle,minimum size=5px,inner sep=0,outer sep=0,fill=tabblu] (d3) at  (0.23-.5+2.2,-0.39+.5) {};
    \node[circle,minimum size=5px,inner sep=0,outer sep=0,fill=tabgre] (d4) at  (0.7 -.5+2.2,-0.71+.5) {};
    \node[circle,minimum size=5px,inner sep=0,outer sep=0,fill=tabgre] (d5) at  (0.64-.5+2.2,-0.35+.5) {};
    
    \node[opacity=.2,transform shape] at (1.1, 0) {\includegraphics[width=1in]{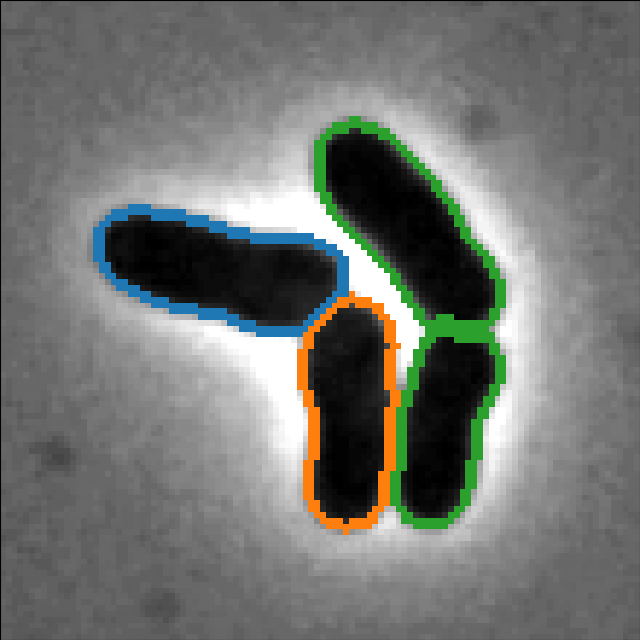}};
    
    \node[transform shape] at(0, 0) {\includegraphics[width=1in]{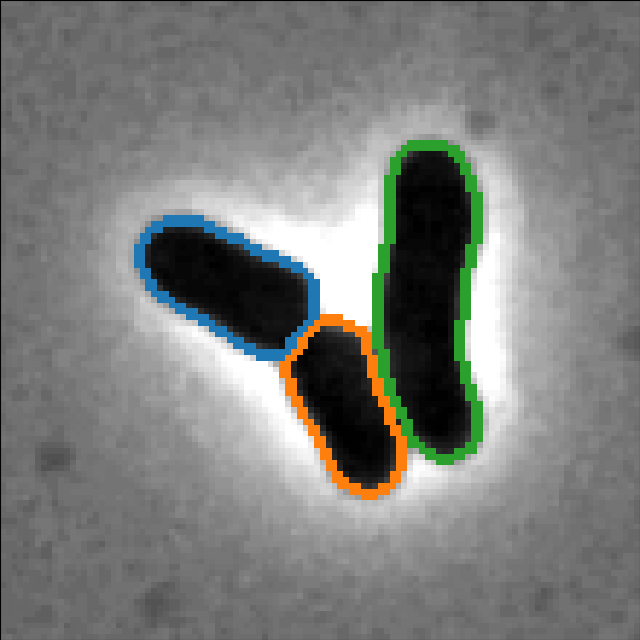}};
    \node[circle,minimum size=5px,inner sep=0,outer sep=0,fill=taborg] (m1) at (0.54-.5,-0.63+.5) {};
    \node[circle,minimum size=5px,inner sep=0,outer sep=0,fill=tabblu] (m2) at (0.35-.5,-0.44+.5) {};
    \node[circle,minimum size=5px,inner sep=0,outer sep=0,fill=tabgre] (m3) at (0.67-.5,-0.47+.5) {};
    
    \node[white] at (0.0-.42, .42) {$t_0$};
    \node[black] at (1.1-.42, .42) {$t_1$};
    \node[white] at (2.2-.42, .42) {$t_2$};

    \draw[red,line width=2pt,{Stealth[length=2mm,width=3mm,round]}-] (d2) -- ++(-.25, -.25);
    
    \def\y{.7};

    \node at (0.975, \y) {or};

    \node at (-0.7, 2*\y) {$A_1$:};
    
    \node[transform shape] at(2.2, 2*\y) {\includegraphics[width=1in]{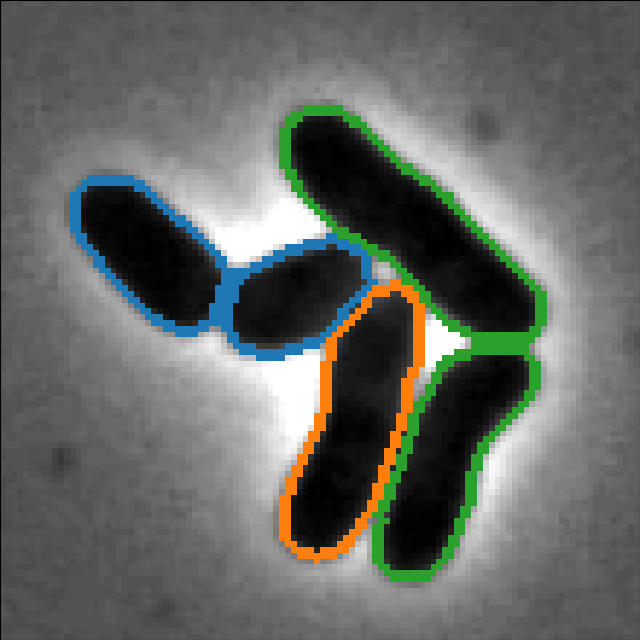}};

    \node[circle,minimum size=5px,inner sep=0,outer sep=0,fill=taborg] (d1) at  (0.55-.5+2.2,-0.66+.5+2*\y) {};
    \node[circle,minimum size=5px,inner sep=0,outer sep=6pt,fill=tabblu] (d2) at  (0.45-.5+2.2,-0.46+.5+2*\y) {};
    \node[circle,minimum size=5px,inner sep=0,outer sep=0,fill=tabblu] (d3) at  (0.23-.5+2.2,-0.39+.5+2*\y) {};
    \node[circle,minimum size=5px,inner sep=0,outer sep=0,fill=tabgre] (d4) at  (0.7 -.5+2.2,-0.71+.5+2*\y) {};
    \node[circle,minimum size=5px,inner sep=0,outer sep=0,fill=tabgre] (d5) at  (0.64-.5+2.2,-0.35+.5+2*\y) {};
    
    \node[opacity=.2,transform shape] at (1.1, 2*\y) {\includegraphics[width=1in]{t1.png}};
    
    \node[transform shape] at(0, 2*\y) {\includegraphics[width=1in]{t0.png}};
    \node[circle,minimum size=5px,inner sep=0,outer sep=0,fill=taborg] (m1) at (0.54-.5,-0.63+.5+2*\y) {};
    \node[circle,minimum size=5px,inner sep=0,outer sep=0,fill=tabblu] (m2) at (0.35-.5,-0.44+.5+2*\y) {};
    \node[circle,minimum size=5px,inner sep=0,outer sep=0,fill=tabgre] (m3) at (0.67-.5,-0.47+.5+2*\y) {};
    
    \node[white] at (0.0-.42, .42+2*\y) {$t_0$};
    \node[black] at (1.1-.42, .42+2*\y) {$t_1$};
    \node[white] at (2.2-.42, .42+2*\y) {$t_2$};

    \draw[red,line width=2pt,{Stealth[length=2mm,width=3mm,round]}-] (d2) -- ++(-.25, -.25);

    \end{tikzpicture}
    }
    \caption{
        Example of ambiguity in two different assignment solutions $A_1$ and $A_2$
        caused by the similar appearance of cells and missing frames.
        Assignments are color-coded, \ie cells in frame $t_2$ with the same color
        as those in $t_0$ are considered daughters of the latter.
        Standard tracking methods yield point estimates, \ie they will choose either 
        solution, but not report any uncertainty caused by the ambiguity in choosing 
        the correct mother for the cell marked by the red arrow.
    }
    \label{fig:tracking-uncertainty}
\end{figure}

\begin{figure*}
    \centering
    
    \begin{subfigure}[t]{.365\textwidth}
    \caption{\makebox[\textwidth]{}}
    \vspace{-3mm}
    \centering
    \resizebox{\columnwidth}{!}{\begin{tikzpicture}[x=1.32in, y=1.32in]
        \node at (-.6, 0) {};
        \node at (0, 0) {\includegraphics[width=1.2in]{t0.png}};
        \node at (1, 0) {\includegraphics[width=1.2in]{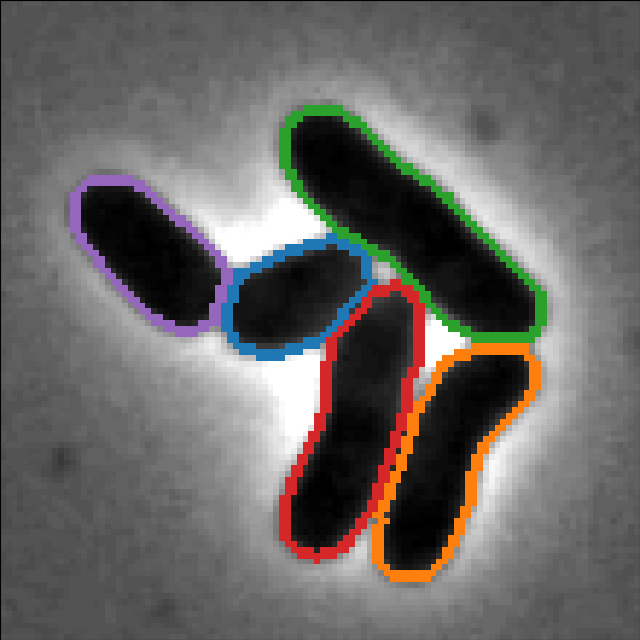}};
        \node[white] at (0-.4, .4) {\large $t$};
        \node[white] at (1-.4, .4) {\large $t'$};

        \node at (0, -.55) {$\cells_t = \{ 
            \begin{tikzpicture}
                \node[circle,minimum size=5px,inner sep=0,outer sep=0,fill=tabblu] {};
            \end{tikzpicture},
            \begin{tikzpicture}
                \node[circle,minimum size=5px,inner sep=0,outer sep=0,fill=taborg] {};
            \end{tikzpicture},
            \begin{tikzpicture}
                \node[circle,minimum size=5px,inner sep=0,outer sep=0,fill=tabgre] {};
            \end{tikzpicture}
        \}$};
        \node at (1, -.55) {$\cells_{t'} = \{ 
            \begin{tikzpicture}
                \node[circle,minimum size=5px,inner sep=0,outer sep=0,fill=tabblu] {};
            \end{tikzpicture},
            \begin{tikzpicture}
                \node[circle,minimum size=5px,inner sep=0,outer sep=0,fill=taborg] {};
            \end{tikzpicture},
            \begin{tikzpicture}
                \node[circle,minimum size=5px,inner sep=0,outer sep=0,fill=tabgre] {};
            \end{tikzpicture},
            \begin{tikzpicture}
                \node[circle,minimum size=5px,inner sep=0,outer sep=0,fill=tabred] {};
            \end{tikzpicture},
            \begin{tikzpicture}
                \node[circle,minimum size=5px,inner sep=0,outer sep=0,fill=tabppl] {};
            \end{tikzpicture}
        \}$};
    \end{tikzpicture}}
    
    \label{fig:imgs}
    \end{subfigure}~\hspace{.025\textwidth}~
    \begin{subfigure}[t]{.23\textwidth}
    \caption{\makebox[\textwidth]{}}
    \vspace{-4mm}
    \centering
    \resizebox{\columnwidth}{!}{\begin{tikzpicture}
        \fill[gray!5]

            -- plot [smooth cycle] coordinates {(-.5,0) (-.2,2.2) (1.2,2.8) (3.2,1.7) (3.4,.2) (2.5, -.3) (1.6,-1)};

        \def\x{0};
        \def\y{0};
        
        \node[circle,minimum size=5px,inner sep=0,outer sep=0,fill=tabppl] (d1) at  (.5+\x,0*.2+\y) {};
        \node[circle,minimum size=5px,inner sep=0,outer sep=0,fill=tabred] (d2) at  (.5+\x,1*.2+\y) {};
        \node[circle,minimum size=5px,inner sep=0,outer sep=0,fill=tabgre] (d3) at  (.5+\x,2*.2+\y) {};
        \node[circle,minimum size=5px,inner sep=0,outer sep=0,fill=taborg] (d4) at  (.5+\x,3*.2+\y) {};
        \node[circle,minimum size=5px,inner sep=0,outer sep=0,fill=tabblu] (d5) at  (.5+\x,4*.2+\y) {};
        \node[circle,minimum size=5px,inner sep=0,outer sep=0,fill=tabgre] (m1) at (0+\x,1*.2+\y) {};
        \node[circle,minimum size=5px,inner sep=0,outer sep=0,fill=taborg] (m2) at (0+\x,2*.2+\y) {};
        \node[circle,minimum size=5px,inner sep=0,outer sep=0,fill=tabblu] (m3) at (0+\x,3*.2+\y) {};
        \draw[line width=1pt] (m1) -- (d1);
        \draw[line width=1pt] (m1) -- (d2);
        \draw[line width=1pt] (m2) -- (d3);
        \draw[line width=1pt] (m3) -- (d4);
        \draw[line width=1pt] (m3) -- (d5);
        
        \def\x{0.3};
        \def\y{1.2};
        
        \node[circle,minimum size=5px,inner sep=0,outer sep=0,fill=tabppl] (d1) at  (.5+\x,0*.2+\y) {};
        \node[circle,minimum size=5px,inner sep=0,outer sep=0,fill=tabred] (d2) at  (.5+\x,1*.2+\y) {};
        \node[circle,minimum size=5px,inner sep=0,outer sep=0,fill=tabgre] (d3) at  (.5+\x,2*.2+\y) {};
        \node[circle,minimum size=5px,inner sep=0,outer sep=0,fill=taborg] (d4) at  (.5+\x,3*.2+\y) {};
        \node[circle,minimum size=5px,inner sep=0,outer sep=0,fill=tabblu] (d5) at  (.5+\x,4*.2+\y) {};
        \node[circle,minimum size=5px,inner sep=0,outer sep=0,fill=tabgre] (m1) at (0+\x,1*.2+\y) {};
        \node[circle,minimum size=5px,inner sep=0,outer sep=0,fill=taborg] (m2) at (0+\x,2*.2+\y) {};
        \node[circle,minimum size=5px,inner sep=0,outer sep=0,fill=tabblu] (m3) at (0+\x,3*.2+\y) {};
        \draw[line width=1pt] (m1) -- (d1);
        \draw[line width=1pt] (m2) -- (d2);
        \draw[line width=1pt] (m2) -- (d3);
        
        \draw[line width=1pt] (m3) -- (d5);
        
        \def\x{1.2};
        \def\y{-0.2};
        
        \node[circle,minimum size=5px,inner sep=0,outer sep=0,fill=tabppl] (d1) at  (.5+\x,0*.2+\y) {};
        \node[circle,minimum size=5px,inner sep=0,outer sep=0,fill=tabred] (d2) at  (.5+\x,1*.2+\y) {};
        \node[circle,minimum size=5px,inner sep=0,outer sep=0,fill=tabgre] (d3) at  (.5+\x,2*.2+\y) {};
        \node[circle,minimum size=5px,inner sep=0,outer sep=0,fill=taborg] (d4) at  (.5+\x,3*.2+\y) {};
        \node[circle,minimum size=5px,inner sep=0,outer sep=0,fill=tabblu] (d5) at  (.5+\x,4*.2+\y) {};
        \node[circle,minimum size=5px,inner sep=0,outer sep=0,fill=tabgre] (m1) at (0+\x,1*.2+\y) {};
        \node[circle,minimum size=5px,inner sep=0,outer sep=0,fill=taborg] (m2) at (0+\x,2*.2+\y) {};
        \node[circle,minimum size=5px,inner sep=0,outer sep=0,fill=tabblu] (m3) at (0+\x,3*.2+\y) {};
        \draw[line width=1pt] (m1) -- (d1);
        \draw[line width=1pt] (m2) -- (d2);
        \draw[line width=1pt] (m3) -- (d3);
        \draw[line width=1pt] (m3) -- (d4);
        \draw[line width=1pt] (m1) -- (d5);
        
        \def\x{1.4};
        \def\y{1.0};
        
        \node[circle,minimum size=5px,inner sep=0,outer sep=0,fill=tabppl] (d1) at  (.5+\x,0*.2+\y) {};
        \node[circle,minimum size=5px,inner sep=0,outer sep=0,fill=tabred] (d2) at  (.5+\x,1*.2+\y) {};
        \node[circle,minimum size=5px,inner sep=0,outer sep=0,fill=tabgre] (d3) at  (.5+\x,2*.2+\y) {};
        \node[circle,minimum size=5px,inner sep=0,outer sep=0,fill=taborg] (d4) at  (.5+\x,3*.2+\y) {};
        \node[circle,minimum size=5px,inner sep=0,outer sep=0,fill=tabblu] (d5) at  (.5+\x,4*.2+\y) {};
        \node[circle,minimum size=5px,inner sep=0,outer sep=0,fill=tabgre] (m1) at (0+\x,1*.2+\y) {};
        \node[circle,minimum size=5px,inner sep=0,outer sep=0,fill=taborg] (m2) at (0+\x,2*.2+\y) {};
        \node[circle,minimum size=5px,inner sep=0,outer sep=0,fill=tabblu] (m3) at (0+\x,3*.2+\y) {};
        \draw[line width=1pt] (m1) -- (d1);
        \draw[line width=1pt] (m2) -- (d2);
        \draw[line width=1pt] (m1) -- (d3);
        \draw[line width=1pt] (m3) -- (d4);
        \draw[line width=1pt] (m3) -- (d5);
        
        \def\x{2.4};
        \def\y{0.4};
        
        \node[circle,minimum size=5px,inner sep=0,outer sep=0,fill=tabppl] (d1) at  (.5+\x,0*.2+\y) {};
        \node[circle,minimum size=5px,inner sep=0,outer sep=0,fill=tabred] (d2) at  (.5+\x,1*.2+\y) {};
        \node[circle,minimum size=5px,inner sep=0,outer sep=0,fill=tabgre] (d3) at  (.5+\x,2*.2+\y) {};
        \node[circle,minimum size=5px,inner sep=0,outer sep=0,fill=taborg] (d4) at  (.5+\x,3*.2+\y) {};
        \node[circle,minimum size=5px,inner sep=0,outer sep=0,fill=tabblu] (d5) at  (.5+\x,4*.2+\y) {};
        \node[circle,minimum size=5px,inner sep=0,outer sep=0,fill=tabgre] (m1) at (0+\x,1*.2+\y) {};
        \node[circle,minimum size=5px,inner sep=0,outer sep=0,fill=taborg] (m2) at (0+\x,2*.2+\y) {};
        \node[circle,minimum size=5px,inner sep=0,outer sep=0,fill=tabblu] (m3) at (0+\x,3*.2+\y) {};
        \draw[line width=1pt] (m1) -- (d1);
        \draw[line width=1pt] (m2) -- (d2);
        \draw[line width=1pt] (m1) -- (d3);
        \draw[line width=1pt] (m2) -- (d4);
        \draw[line width=1pt] (m3) -- (d5);

        \def\x{3.4};
        \def\y{1.7};
        
        \node[opacity=.3,circle,minimum size=5px,inner sep=0,outer sep=0,fill=tabppl] (d1) at  (.5+\x,0*.2+\y) {};
        \node[opacity=.3,circle,minimum size=5px,inner sep=0,outer sep=0,fill=tabred] (d2) at  (.5+\x,1*.2+\y) {};
        \node[opacity=.3,circle,minimum size=5px,inner sep=0,outer sep=0,fill=tabgre] (d3) at  (.5+\x,2*.2+\y) {};
        \node[opacity=.3,circle,minimum size=5px,inner sep=0,outer sep=0,fill=taborg] (d4) at  (.5+\x,3*.2+\y) {};
        \node[opacity=.3,circle,minimum size=5px,inner sep=0,outer sep=0,fill=tabblu] (d5) at  (.5+\x,4*.2+\y) {};
        \node[opacity=.3,circle,minimum size=5px,inner sep=0,outer sep=0,fill=tabgre] (m1) at (0+\x,1*.2+\y) {};
        \node[opacity=.3,circle,minimum size=5px,inner sep=0,outer sep=0,fill=taborg] (m2) at (0+\x,2*.2+\y) {};
        \node[opacity=.3,circle,minimum size=5px,inner sep=0,outer sep=0,fill=tabblu] (m3) at (0+\x,3*.2+\y) {};
        \draw[opacity=.3,line width=1pt] (m1) -- (d1);
        \draw[opacity=.3,line width=1pt] (m2) -- (d2);
        \draw[opacity=.3,line width=1pt] (m1) -- (d3);
        \draw[opacity=.3,line width=1pt] (m2) -- (d4);
        \draw[opacity=.3,line width=1pt] (m3) -- (d5);
        \draw[opacity=.3,line width=1pt] (m3) -- (d4);

        \def\x{-1.5};
        \def\y{0.4};
        
        \node[opacity=.3,circle,minimum size=5px,inner sep=0,outer sep=0,fill=tabppl] (d1) at  (.5+\x,0*.2+\y) {};
        \node[opacity=.3,circle,minimum size=5px,inner sep=0,outer sep=0,fill=tabred] (d2) at  (.5+\x,1*.2+\y) {};
        \node[opacity=.3,circle,minimum size=5px,inner sep=0,outer sep=0,fill=tabgre] (d3) at  (.5+\x,2*.2+\y) {};
        \node[opacity=.3,circle,minimum size=5px,inner sep=0,outer sep=0,fill=taborg] (d4) at  (.5+\x,3*.2+\y) {};
        \node[opacity=.3,circle,minimum size=5px,inner sep=0,outer sep=0,fill=tabblu] (d5) at  (.5+\x,4*.2+\y) {};
        \node[opacity=.3,circle,minimum size=5px,inner sep=0,outer sep=0,fill=tabgre] (m1) at (0+\x,1*.2+\y) {};
        \node[opacity=.3,circle,minimum size=5px,inner sep=0,outer sep=0,fill=taborg] (m2) at (0+\x,2*.2+\y) {};
        \node[opacity=.3,circle,minimum size=5px,inner sep=0,outer sep=0,fill=tabblu] (m3) at (0+\x,3*.2+\y) {};
        \draw[opacity=.3,line width=1pt] (m1) -- (d1);
        \draw[opacity=.3,line width=1pt] (m2) -- (d2);
        \draw[opacity=.3,line width=1pt] (m1) -- (d3);
        \draw[opacity=.3,line width=1pt] (m2) -- (d4);
        \draw[opacity=.3,line width=1pt] (m3) -- (d5);
        \draw[opacity=.3,line width=1pt] (m3) -- (d3);

        \def\x{2.8};
        \def\y{-1.2};
        
        \node[opacity=.3,circle,minimum size=5px,inner sep=0,outer sep=0,fill=tabppl] (d1) at  (.5+\x,0*.2+\y) {};
        \node[opacity=.3,circle,minimum size=5px,inner sep=0,outer sep=0,fill=tabred] (d2) at  (.5+\x,1*.2+\y) {};
        \node[opacity=.3,circle,minimum size=5px,inner sep=0,outer sep=0,fill=tabgre] (d3) at  (.5+\x,2*.2+\y) {};
        \node[opacity=.3,circle,minimum size=5px,inner sep=0,outer sep=0,fill=taborg] (d4) at  (.5+\x,3*.2+\y) {};
        \node[opacity=.3,circle,minimum size=5px,inner sep=0,outer sep=0,fill=tabblu] (d5) at  (.5+\x,4*.2+\y) {};
        \node[opacity=.3,circle,minimum size=5px,inner sep=0,outer sep=0,fill=tabgre] (m1) at (0+\x,1*.2+\y) {};
        \node[opacity=.3,circle,minimum size=5px,inner sep=0,outer sep=0,fill=taborg] (m2) at (0+\x,2*.2+\y) {};
        \node[opacity=.3,circle,minimum size=5px,inner sep=0,outer sep=0,fill=tabblu] (m3) at (0+\x,3*.2+\y) {};
        \draw[opacity=.3,line width=1pt] (m1) -- (d1);
        \draw[opacity=.3,line width=1pt] (m1) -- (d2);
        \draw[opacity=.3,line width=1pt] (m2) -- (d3);
        \draw[opacity=.3,line width=1pt] (m2) -- (d4);
        \draw[opacity=.3,line width=1pt] (m2) -- (d5);
        \draw[opacity=.3,line width=1pt] (m3) -- (d5);

        \draw plot [smooth cycle] coordinates {(-.5,0) (-.2,2.2) (1.2,2.8) (3.2,1.7) (3.4,.2) (2.5, -.3) (1.6,-1)};
        \node at (1.2, 2.5) {$\assignments$};
        \node at (-.5, -1) {$\mathcal{P}(\cells_t \times \cells_{t'})$};

        \node at (2, -1.5) {};
    \end{tikzpicture}}
    \label{fig:prior}
    \end{subfigure}~\hspace{.025\textwidth}~
    \begin{subfigure}[t]{.35\textwidth}
    \caption{\makebox[.85\textwidth]{}}
    \vspace{-3mm}
    \centering
    \resizebox{\columnwidth}{!}{\begin{tikzpicture}
    \begin{scope}[cm={.9,0,0,1.1,(.5,0)}]
        \def\basex{5.5};
        \def\basey{-.5};

        \def\x{\basex};
        \def\y{\basey};
        
        \draw[->] (0+\x,0+\y) -- (0+\x,3+\y);
        \draw[->] (0+\x,0+\y) -- (5.2+\x,0+\y);
        
        \def\x{\basex+0.35};
        \def\y{\basey-0.60};
        
        \node[circle,minimum size=3px,inner sep=0,outer sep=0,fill=tabppl] (d1) at (.3+\x,0*.12+\y) {};
        \node[circle,minimum size=3px,inner sep=0,outer sep=0,fill=tabred] (d2) at (.3+\x,1*.12+\y) {};
        \node[circle,minimum size=3px,inner sep=0,outer sep=0,fill=tabgre] (d3) at (.3+\x,2*.12+\y) {};
        \node[circle,minimum size=3px,inner sep=0,outer sep=0,fill=taborg] (d4) at (.3+\x,3*.12+\y) {};
        \node[circle,minimum size=3px,inner sep=0,outer sep=0,fill=tabblu] (d5) at (.3+\x,4*.12+\y) {};
        \node[circle,minimum size=3px,inner sep=0,outer sep=0,fill=tabgre] (m1) at (.0+\x,1*.12+\y) {};
        \node[circle,minimum size=3px,inner sep=0,outer sep=0,fill=taborg] (m2) at (.0+\x,2*.12+\y) {};
        \node[circle,minimum size=3px,inner sep=0,outer sep=0,fill=tabblu] (m3) at (.0+\x,3*.12+\y) {};
        \draw[line width=.6pt] (m1) -- (d1);
        \draw[line width=.6pt] (m1) -- (d2);
        \draw[line width=.6pt] (m2) -- (d3);
        \draw[line width=.6pt] (m3) -- (d4);
        \draw[line width=.6pt] (m3) -- (d5);
        
        \def\x{\basex+1.35};
        
        \node[circle,minimum size=3px,inner sep=0,outer sep=0,fill=tabppl] (d1) at (.3+\x,0*.12+\y) {};
        \node[circle,minimum size=3px,inner sep=0,outer sep=0,fill=tabred] (d2) at (.3+\x,1*.12+\y) {};
        \node[circle,minimum size=3px,inner sep=0,outer sep=0,fill=tabgre] (d3) at (.3+\x,2*.12+\y) {};
        \node[circle,minimum size=3px,inner sep=0,outer sep=0,fill=taborg] (d4) at (.3+\x,3*.12+\y) {};
        \node[circle,minimum size=3px,inner sep=0,outer sep=0,fill=tabblu] (d5) at (.3+\x,4*.12+\y) {};
        \node[circle,minimum size=3px,inner sep=0,outer sep=0,fill=tabgre] (m1) at (.0+\x,1*.12+\y) {};
        \node[circle,minimum size=3px,inner sep=0,outer sep=0,fill=taborg] (m2) at (.0+\x,2*.12+\y) {};
        \node[circle,minimum size=3px,inner sep=0,outer sep=0,fill=tabblu] (m3) at (.0+\x,3*.12+\y) {};
        \draw[line width=.6pt] (m1) -- (d1);
        \draw[line width=.6pt] (m2) -- (d2);
        \draw[line width=.6pt] (m2) -- (d3);
        \draw[line width=.6pt] (m3) -- (d4);
        \draw[line width=.6pt] (m3) -- (d5);
        
        \def\x{\basex+2.35};
        
        \node[circle,minimum size=3px,inner sep=0,outer sep=0,fill=tabppl] (d1) at (.3+\x,0*.12+\y) {};
        \node[circle,minimum size=3px,inner sep=0,outer sep=0,fill=tabred] (d2) at (.3+\x,1*.12+\y) {};
        \node[circle,minimum size=3px,inner sep=0,outer sep=0,fill=tabgre] (d3) at (.3+\x,2*.12+\y) {};
        \node[circle,minimum size=3px,inner sep=0,outer sep=0,fill=taborg] (d4) at (.3+\x,3*.12+\y) {};
        \node[circle,minimum size=3px,inner sep=0,outer sep=0,fill=tabblu] (d5) at (.3+\x,4*.12+\y) {};
        \node[circle,minimum size=3px,inner sep=0,outer sep=0,fill=tabgre] (m1) at (.0+\x,1*.12+\y) {};
        \node[circle,minimum size=3px,inner sep=0,outer sep=0,fill=taborg] (m2) at (.0+\x,2*.12+\y) {};
        \node[circle,minimum size=3px,inner sep=0,outer sep=0,fill=tabblu] (m3) at (.0+\x,3*.12+\y) {};
        \draw[line width=.6pt] (m1) -- (d1);
        \draw[line width=.6pt] (m2) -- (d2);
        \draw[line width=.6pt] (m3) -- (d3);
        \draw[line width=.6pt] (m3) -- (d4);
        \draw[line width=.6pt] (m1) -- (d5);
        
        \def\x{\basex+3.35};
        
        \node[circle,minimum size=3px,inner sep=0,outer sep=0,fill=tabppl] (d1) at (.3+\x,0*.12+\y) {};
        \node[circle,minimum size=3px,inner sep=0,outer sep=0,fill=tabred] (d2) at (.3+\x,1*.12+\y) {};
        \node[circle,minimum size=3px,inner sep=0,outer sep=0,fill=tabgre] (d3) at (.3+\x,2*.12+\y) {};
        \node[circle,minimum size=3px,inner sep=0,outer sep=0,fill=taborg] (d4) at (.3+\x,3*.12+\y) {};
        \node[circle,minimum size=3px,inner sep=0,outer sep=0,fill=tabblu] (d5) at (.3+\x,4*.12+\y) {};
        \node[circle,minimum size=3px,inner sep=0,outer sep=0,fill=tabgre] (m1) at (.0+\x,1*.12+\y) {};
        \node[circle,minimum size=3px,inner sep=0,outer sep=0,fill=taborg] (m2) at (.0+\x,2*.12+\y) {};
        \node[circle,minimum size=3px,inner sep=0,outer sep=0,fill=tabblu] (m3) at (.0+\x,3*.12+\y) {};
        \draw[line width=.6pt] (m1) -- (d1);
        \draw[line width=.6pt] (m2) -- (d2);
        \draw[line width=.6pt] (m1) -- (d3);
        \draw[line width=.6pt] (m3) -- (d4);
        \draw[line width=.6pt] (m3) -- (d5);
        
        \def\x{\basex+4.35};
        
        \node[circle,minimum size=3px,inner sep=0,outer sep=0,fill=tabppl] (d1) at (.3+\x,0*.12+\y) {};
        \node[circle,minimum size=3px,inner sep=0,outer sep=0,fill=tabred] (d2) at (.3+\x,1*.12+\y) {};
        \node[circle,minimum size=3px,inner sep=0,outer sep=0,fill=tabgre] (d3) at (.3+\x,2*.12+\y) {};
        \node[circle,minimum size=3px,inner sep=0,outer sep=0,fill=taborg] (d4) at (.3+\x,3*.12+\y) {};
        \node[circle,minimum size=3px,inner sep=0,outer sep=0,fill=tabblu] (d5) at (.3+\x,4*.12+\y) {};
        \node[circle,minimum size=3px,inner sep=0,outer sep=0,fill=tabgre] (m1) at (.0+\x,1*.12+\y) {};
        \node[circle,minimum size=3px,inner sep=0,outer sep=0,fill=taborg] (m2) at (.0+\x,2*.12+\y) {};
        \node[circle,minimum size=3px,inner sep=0,outer sep=0,fill=tabblu] (m3) at (.0+\x,3*.12+\y) {};
        \draw[line width=.6pt] (m1) -- (d1);
        \draw[line width=.6pt] (m2) -- (d2);
        \draw[line width=.6pt] (m1) -- (d3);
        \draw[line width=.6pt] (m2) -- (d4);
        \draw[line width=.6pt] (m3) -- (d5);

        \def\x{\basex};
        \def\y{\basey};

        \node at (\x+5.2, \y-.36) {$\assignments$};

        \draw[line width=2pt,tabblu] plot coordinates {(\x+2.0, \y+2.5+.2) (\x+2.4, \y+2.5+.2)};
        \draw[line width=2pt,taborg] plot coordinates {(\x+2.0, \y+2.0+.2) (\x+2.4, \y+2.0+.2)};
        
        \node[anchor=west,align=left] at (\x+2.4, \y+2.5+.2) {\small $\prob(\assignment | \cells_{t}, \cells_{t'})$};
        \node[anchor=west,align=left] at (\x+2.4, \y+2.0+.2) {\small $\{p_k\}_{k=1}^K$, Eq.~\eqref{eq:sni}};

        \draw[line width=2pt,taborg,fill=taborg!10] plot coordinates {
            (\x+.1  , \y) 
            (\x+.1  , \y+.1+2.5) 
            (\x+.1+1, \y+.1+2.5) 
            (\x+.1+1, \y+.1+1.6) 
            (\x+.1+2, \y+.1+1.6) 
            (\x+.1+2, \y+.1+1) 
            (\x+.1+3, \y+.1+1) 
            (\x+.1+3, \y)  };

        \draw[line width=2pt,tabblu] plot coordinates {
            (\x+.05  , \y) 
            (\x+.05  , \y+2.5) 
            (\x+.05+1, \y+2.5) 
            (\x+.05+1, \y+1.6) 
            (\x+.05+2, \y+1.6) 
            (\x+.05+2, \y+1) 
            (\x+.05+3, \y+1) 
            (\x+.05+3, \y+0.8) 
            (\x+.05+4, \y+0.8) 
            (\x+.05+4, \y+0.4) 
            (\x+.05+5, \y+0.4) };

        \fill[tabblu,opacity=.1] plot coordinates {
            (\x+.05  , \y) 
            (\x+.05  , \y+2.5) 
            (\x+.05+1, \y+2.5) 
            (\x+.05+1, \y+1.6) 
            (\x+.05+2, \y+1.6) 
            (\x+.05+2, \y+1) 
            (\x+.05+3, \y+1) 
            (\x+.05+3, \y+0.8) 
            (\x+.05+4, \y+0.8) 
            (\x+.05+4, \y+0.4) 
            (\x+.05+5, \y+0.4)
            (\x+.05+5, \y) };
        \end{scope}
        
        \node at (10.75, .75) {\scalebox{1.5}{$\to$}};
        \node[align=center] at (12, 2.25) {
            \small Edge proba- \\[-1mm] 
            \small bilities $P$, \\[-1mm] 
            \small Eq.~\eqref{eq:importance-weighted}};
        \begin{scope}[cm={3, 0, 0, 3, (12, .75)}]
            \node[circle,minimum size=7px,inner sep=0,outer sep=0,fill=tabppl] (d1) at ( .15,-2*.12) {};
            \node[circle,minimum size=7px,inner sep=0,outer sep=0,fill=tabred] (d2) at ( .15,-1*.12) {};
            \node[circle,minimum size=7px,inner sep=0,outer sep=0,fill=tabgre] (d3) at ( .15, 0*.12) {};
            \node[circle,minimum size=7px,inner sep=0,outer sep=0,fill=taborg] (d4) at ( .15, 1*.12) {};
            \node[circle,minimum size=7px,inner sep=0,outer sep=0,fill=tabblu] (d5) at ( .15, 2*.12) {};
            \node[circle,minimum size=7px,inner sep=0,outer sep=0,fill=tabgre] (m1) at (-.15,-1*.12) {};
            \node[circle,minimum size=7px,inner sep=0,outer sep=0,fill=taborg] (m2) at (-.15, 0*.12) {};
            \node[circle,minimum size=7px,inner sep=0,outer sep=0,fill=tabblu] (m3) at (-.15, 1*.12) {};

            \draw[line width=1.5pt,black!10] (m3) -- (d5);
            \draw[line width=1.5pt,black!10] (m2) -- (d5);
            \draw[line width=1.5pt,black!10] (m1) -- (d5);
            
            \draw[line width=1.5pt,black!10] (m1) -- (d4);

            \draw[line width=1.5pt,black!45] (m2) -- (d4);

            \draw[line width=1.5pt] (m1) -- (d3);
            \draw[line width=1.5pt] (m2) -- (d2);
            \draw[line width=1.5pt] (m3) -- (d1);
        \end{scope}
    \end{tikzpicture}}
    \label{fig:sni}
    \end{subfigure}
    \vspace{-4mm}
    \caption{
        Schematic presentation of our Bayesian perspective and inference method
        for uncertainty-aware cell tracking.
        \subref{fig:imgs} depicts the detections $\cells_t, \cells_{t'}$ of two 
        consecutive frames, which are the input to tracking methods from the TbD 
        paradigm.
        \subref{fig:prior} The set of biologically feasible assignments $\assignments$
        forms a subset within the set of all possible many-to-many assignments.
        \subref{fig:sni} Sorting assignment solutions by their posterior density
        we approximate the full posterior by a set of the $K$ most plausible
        solutions. 
        Edge probabilities are estimated as self-normalized importance-weighted
        average (\cf \Cref{eq:sni}). 
    }
    \label{fig:bayes}
\end{figure*}

The strong reliance on computational tools demands for high reliability and
trustworthiness, which is improved by uncertainty-aware analyses aiming to 
reliably estimate the confidence in their own predictions.
While over-confidence in deep neural networks encountering distribution 
shifts is a commonly known issue in the deep
learning community \cite{guo_calibration_2017,kristiadi_being_2020}, 
uncertainty estimation as a remedy has -- so far -- attracted only little 
attention in cell tracking.

In this paper, we strive to make cell tracking more reliable by complementing it 
with uncertainty estimation.
To this end, we explore two perspectives on the tracking problem, considering
it as a Bayesian inference and as a classification problem. 
Our Bayesian perspective gives rise to the \emph{cell tracking posterior},
motivating sampling and approximate sampling methods for uncertainty quantification,
which however come at increased computational costs.
The classification perspective provides less costly, nevertheless useful alternatives
to quantify uncertainty.
Moreover, the classification perspective also provides tools to evaluate and calibrate
uncertainty estimates using known techniques such as \emph{temperature scaling} 
\cite{guo_calibration_2017}.
Notably, the methods under consideration are applicable 
to the large family of linear assignment-based tracking methods, which are embedded naturally within our framework.
We demonstrate this by testing our methods 
on various tracking 
algorithms, namely distance- \cite{crocker_methods_1996, fukai_laptrack_2023}, 
overlap- \cite{fukai_laptrack_2023}, and activity-based cell tracking 
\cite{ruzaeva_cell_2022}
as well as a novel, recently presented Transformer-based cell tracking algorithm
\cite{gallusser_trackastra_2024}.
Our results show that, indeed, vanilla cell tracking algorithms tend to be 
overconfident, especially at low temporal resolution,
though their uncertainties still prove useful in that they positively
correlate with the tracking performance.
Moreover, we show that the methods developed in \Cref{sec:framework} improve 
the calibration of uncertainty estimation 
in cell tracking.
Most notably, temperature scaling \cite{guo_calibration_2017} turns out
to be a cheap, but useful method to greatly improve calibration of tracking
uncertainty if ground truth tracking data is available.

\section{Probabilistic Cell Tracking}
\label{sec:framework}

A major contributor to uncertainty in cell tracking are low frame rates at 
which microscopy images are often acquired (cf. \Cref{fig:tracking-uncertainty}).
The frame rate is often limited by technical or biological factors, 
like \eg camera and stage movement speed \cite{seiffarth_tracking_2024} 
or phototoxicity \cite{tinevez_chapter_2012}.
For fast growing organisms, especially microbes,
this becomes an issue when the dynamics within the population become
too fast to allow for unambiguous tracking \cite{theorell_when_2019, paul_robust_2024}.
We sketch the issue by means of an illustrative example in \Cref{fig:tracking-uncertainty},
where one of the daughter cells (indicated by red arrows) is located such that
a solely position-based association to the mother cell becomes ambiguous.
In such a case, an off-the-shelf assignment solver will yield either of the
two solutions, possibly choosing the wrong one, but remaining unaware of the alternative solution,
and hence being unreliable.
An uncertainty-aware cell tracking algorithm on the other hand should be able to 
express its uncertainty about the generated solution,
\eg in terms of a confidence score, 
indicating the ambiguity of the given problem.
In the following, 
we introduce and discuss techniques 
for estimating the uncertainty in cell tracking.

\subsection{Cell Tracking as Bayesian Inference Problem}
\label{sec:bayesian-inference}

Our consideration of the cell tracking problem starts with
two consecutive frames of a microscopy image sequence with detections $\cells_t, \cells_{t'}$ at times 
$t < t'$ such as exemplarily 
shown in \Cref{fig:imgs}.
At the core of most cell tracking algorithm lies the choice of a cost function 
$w: \cells_t \times \cells_{t'} \to \R_+$.
Assuming $\cell_i, \cell_j'$ denote the positions of the cells in frame $t$ and $t'$, respectively, 
\ie $\cell_i, \cell_j' \in \R^d$ with $d=2$ or $d=3$ typically,
the commonly used cost function 
\begin{align}
    \label{eq:cost}
    w_{_\text{L2}}(\cell_i, \cell_j') = \frac{\lambda}{2} \| \cell_i - \cell_j' \|_2^2
\end{align}
is equivalent to assuming that cells behave in a Brownian fashion \cite{crocker_methods_1996}, \ie
\begin{align}
    \cell_j' = \cell_i + \, \epsilon, \ 
        \text{ for } \epsilon \sim \normal( 0, \lambda^{-1}I)
\end{align}
with $\lambda$ being a hyperparameter representing the movement speed.
From this we 
derive the likelihood $\prob(\cell_j' \cond \cell_i)$ of
observing cell $\cell_i$ from the frame at time $t$ at the position $\cell_j'$ in 
the next frame at time $t'$.

Recently, cost functions have also been implemented by means of 
neural networks \cite{gallusser_trackastra_2024, ben-haim_graph_2022},
allowing to learn them from data rather than handcrafting them.
Many of those techniques follow a simple paradigm, where a neural network is
used as a feature extractor $\nn: \cells \to \R^d$ with learnable parameters $\theta$,
that maps arbitrary cell detections and its features $\cells$
to some continuous latent space \cite{gallusser_trackastra_2024, ben-haim_graph_2022}.
The neural network-based cost function is then computed as
\begin{align}
    \label{eq:nn-cost}
    w_{_\text{NN}}(\cell_i, \cell_j') = \frac{1}{2} \| \nn(\cell_i) - \nn(\cell_j') \|_2^2.
\end{align}

In the case of multiple cells $\cell_1, \ldots, \cell_m \in \cells_t$ and 
$\cell_1', \ldots, \cell_n' \in \cells_{t'}$, we aim to find an \emph{assignment}
$\assignment \in \mathcal{P}(\cells_t \times \cells_{t'})$,
such that every mother cell is represented at most twice
and every daughter cell is represented at most once.
We denote the subset of all assignments that adhere to those 
constraints as $\assignments \subset \mathcal{P}(\cells_t \times \cells_{t'})$, 
the \emph{biologically feasible assignments} (\cf \Cref{fig:prior}).
The common approach to solve for a single assignment
is to optimize
the joint likelihood of observing cells $\cells_{t'}$ given $\cells_t$ and an assignment $\assignment$,
\ie
\begin{align}
    \label{eq:optimize}
    \optassignment 
    &
    := \argmax_{\assignment \in \assignments} \
        \prob(\cells_{t'} \cond \cells_t, \assignment) 
    \\
    \label{eq:ilp_obj}
    &
    \phantom{:}= \argmax_{\assignment \in \assignments} 
        \sum_{(\cell_i, \cell_j') \in \assignment} \hspace{-3mm} -w(\cell_i, \cell_j') - m w_a - n w_d
\end{align}
where $m$ and $n$ are the numbers of appearing and disappearing cells, 
$-w_a$ and $-w_d$ are the respective log probabilities of the events,
and $w(\cell_i, \cell_j')$ is some chosen cost function like those from 
\Cref{eq:cost} \& \eqref{eq:nn-cost} or \Cref{app:cost-functions}.
A detailed derivation of \Cref{eq:optimize} \& \eqref{eq:ilp_obj} is given in \Cref{app:ilp}.

Conveniently, \Cref{eq:ilp_obj} is the solution of a linear assignment problem (LAP).
Considering that the structure of the solution $\assignment \in \assignments$,
\ie its biological feasibility,
is known \emph{a priori} without the need for any actual observations,
the formulation from \Cref{eq:optimize} lends itself nicely to a Bayesian interpretation
with $\optassignment$ being the \emph{maximum a posteriori} (MAP) estimate of the posterior distribution~of~assignments~
\begin{align}
    \label{eq:posterior}
    \prob(\assignment \cond \cells_{t'}, \cells_t) 
    = \frac {\prob(\cells_{t'} \cond \cells_t, \assignment) \prob(\assignment)}
            {\sum_{A \in \mathcal{A}} \prob(\cells_{t'} \cond \cells_t, \assignment) \prob(\assignment)},
\end{align}
where we choose the uniform distribution over valid assignments $\assignments$ as prior distribution, 
$\prob(\assignment) := \flatfrac{\ind{ \assignment \in \assignments}}{|\assignments|}$.
Given this \emph{cell tracking posterior} distribution,
we obtain the predictive distribution of the event that $\cell_i$ is the mother of~$\cell_j'$
\begin{align}
    \label{eq:predictive-posterior}
    P_{ij} &:= \prob((\cell_i, \cell_j') \in \assignment \cond \cells_{t}, \cells_{t'}) \\
        &= \sum_{\assignment \in \assignments} \ind{(\cell_i, \cell_j') \in \assignment} 
            \prob(\assignment \cond \cells_{t}, \cells_{t'})
\end{align}
as the weighted frequency of $\cell_i$ being the mother of $\cell_j'$ among all possible solutions,
where we drop the explicit dependence on the $t$-th and $t'$-th frame for brevity in $P_{ij}$.
Based on the close connection between the LAP and that of finding
maximal weight matchings in bipartite graphs, 
we refer to a daughter-mother pair $(\cell_i, \cell_j')$
as an \emph{edge} (\cf \Cref{fig:bayes}).

\subsubsection{Bayesian Inference on Assignments}
\label{sec:bayesian-assignment}

The Bayesian formulation of cell tracking given in \Cref{eq:posterior}
suggests to sample said posterior instead of just obtaining a single solution
maximizing its density.
This idea is also known as \emph{joint probabilistic data association} 
\cite{fortmann_sonar_1983}.
Given a set of $K$ assignments 
$\assignment_1, \ldots, \assignment_K \sim \prob(\assignment \cond \cells_t, \cells_{t'})$
sampled from the posterior distribution in \Cref{eq:posterior},
we may estimate the predictive posterior from \Cref{eq:predictive-posterior} as~
\begin{align}
    \label{eq:monte-carlo}
    P_{ij}
        \approx \frac{1}{K} \sum_{k=1}^K \ind{(\cell_i, \cell_j') \in \assignment_k}.
\end{align}
Few works have considered the arising sampling problem thus far within
the context of cell tracking.
The close connection of solving \Cref{eq:ilp_obj} and bipartite maximum weight
matching observed in \cite{ruzaeva_cell_2022} 
suggests to use 
Markov chain Monte Carlo algorithms for sampling assignments.
Alternatively, as $\assignments$ is a finite set, one can solve 
\Cref{eq:predictive-posterior}
exactly by means of enumerating the solution set $\assignments$.
The size of $\assignments$, however, is at least 
$\mathcal{O}(N!)$ with $N=\min\{|\cells_{t}|, |\cells_{t'}|\}$.
Nevertheless, enumeration of the top-$K$ best solutions to an LAP
is possible \cite{rezatofighi_joint_2015} and offered by commercial LAP 
solvers like \gurobi.
This allows to estimate \Cref{eq:predictive-posterior} 
by means of the self-normalized importance-weighted estimator \cite{tokdar_importance_2010}~
\begin{align}
    \label{eq:importance-weighted}
    P_{ij}
        \approx \sum_{k=1}^{K} \ind{(\cell_i, \cell_j') \in \assignment_k} \cdot p_k,
\end{align}
with weight~
\begin{align}
    \label{eq:sni}
    p_k := \frac{\prob(\cells_{t'} \cond \cells_t, \assignment_k) \prob(\assignment_k)}
            {\sum_{l=1}^{K} \prob(\cells_{t'} \cond \cells_t, \assignment_l) \prob(\assignment_l)}, 
\end{align}
for $k=1, \ldots, K$ and where we now assumed $\assignment_1, \ldots, \assignment_{K}$ 
to be the top-$K$ best solutions.
We sketch this method in \Cref{fig:sni}.

\subsubsection{Feature Perturbations}
\label{sec:feature-perturbation}

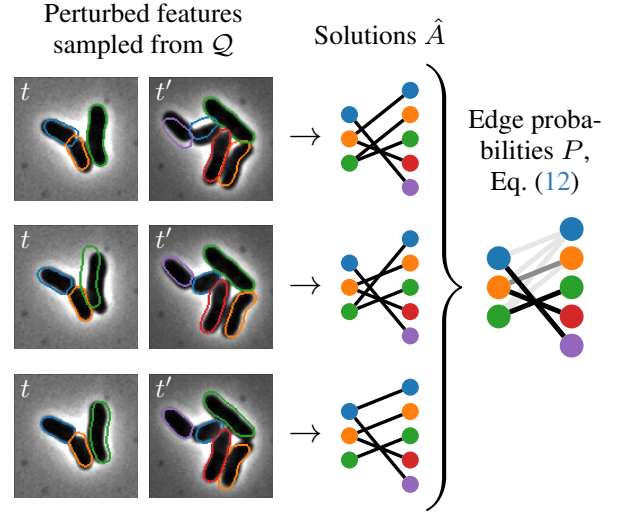
\begin{figure}
    \resizebox{1\columnwidth}{!}{
        \begin{tikzpicture}[x=.55in, y=.55in]
            \node[align=center] at (.5, 3) {\footnotesize Perturbed features \\[-1mm] \footnotesize sampled from $\mathcal{Q}$};
            \node[align=center] at (2.25, 3) {\footnotesize Solutions $\hat{A}$};

            \node at (0.0, 0.0) {\includegraphics[width=.5in]{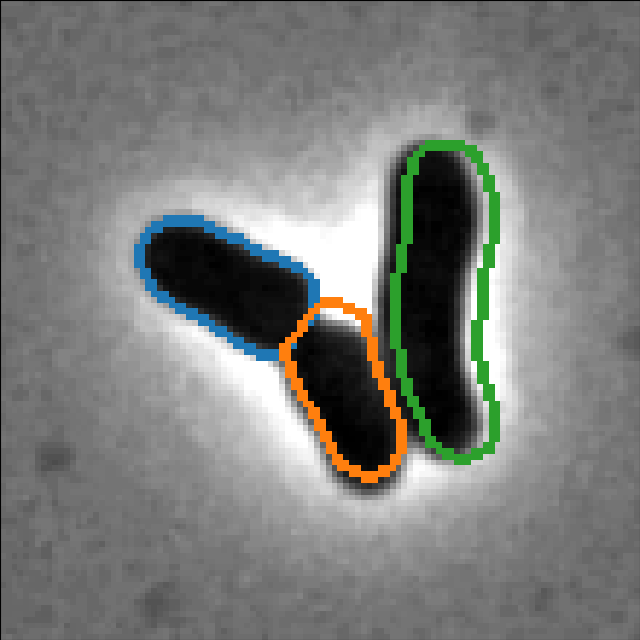}};
            \node at (1.0, 0.0) {\includegraphics[width=.5in]{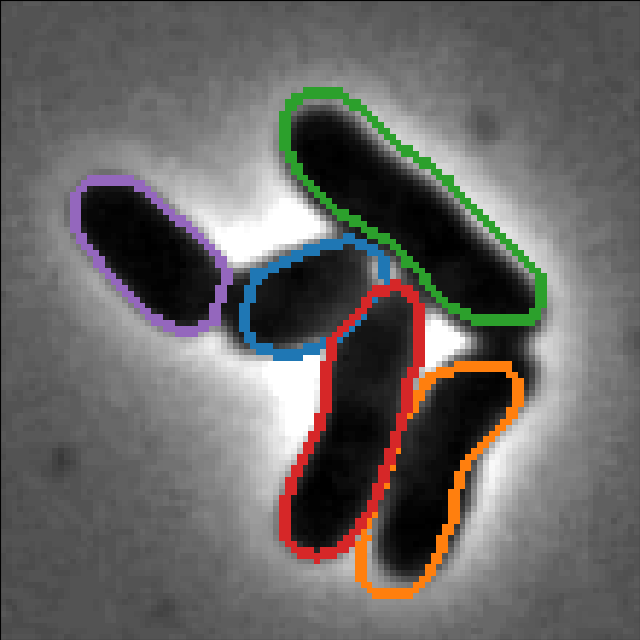}};
            \node[anchor=south west,white] at (0-.5, 0.2) {\scalebox{.8}{$t$}};
            \node[anchor=south west,white] at (1-.5, 0.2) {\scalebox{.8}{$t'$}};
            \node at (1.7, 0.0) {$\to$};
            
            \begin{scope}[cm={1.5, 0, 0, 1.5, (2.25, 0)}]
                \node[circle,minimum size=5px,inner sep=0,outer sep=0,fill=tabppl] (d1) at ( .15,-2*.12) {};
                \node[circle,minimum size=5px,inner sep=0,outer sep=0,fill=tabred] (d2) at ( .15,-1*.12) {};
                \node[circle,minimum size=5px,inner sep=0,outer sep=0,fill=tabgre] (d3) at ( .15, 0*.12) {};
                \node[circle,minimum size=5px,inner sep=0,outer sep=0,fill=taborg] (d4) at ( .15, 1*.12) {};
                \node[circle,minimum size=5px,inner sep=0,outer sep=0,fill=tabblu] (d5) at ( .15, 2*.12) {};
                \node[circle,minimum size=5px,inner sep=0,outer sep=0,fill=tabgre] (m1) at (-.15,-1*.12) {};
                \node[circle,minimum size=5px,inner sep=0,outer sep=0,fill=taborg] (m2) at (-.15, 0*.12) {};
                \node[circle,minimum size=5px,inner sep=0,outer sep=0,fill=tabblu] (m3) at (-.15, 1*.12) {};
                \draw[line width=1pt] (m1) -- (d3);
                \draw[line width=1pt] (m2) -- (d4);
                \draw[line width=1pt] (m2) -- (d2);
                \draw[line width=1pt] (m3) -- (d1);
                \draw[line width=1pt] (m3) -- (d5);
            \end{scope}
        
            \node at (0.0, 1.1) {\includegraphics[width=.5in]{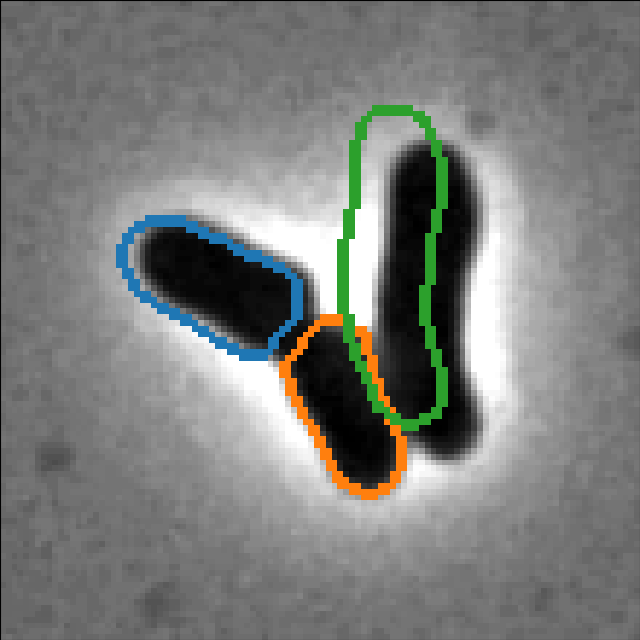}};
            \node at (1.0, 1.1) {\includegraphics[width=.5in]{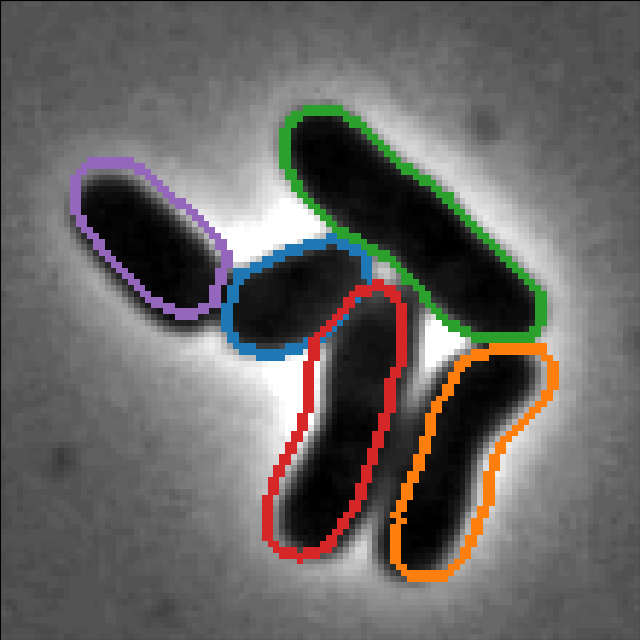}};
            \node[anchor=south west,white] at (0-.5, 1.3) {\scalebox{.8}{$t$}};
            \node[anchor=south west,white] at (1-.5, 1.3) {\scalebox{.8}{$t'$}};
            \node at (1.7, 1.1) {$\to$};
            
            \begin{scope}[cm={1.5, 0, 0, 1.5, (2.25, 1.1)}]
                \node[circle,minimum size=5px,inner sep=0,outer sep=0,fill=tabppl] (d1) at ( .15,-2*.12) {};
                \node[circle,minimum size=5px,inner sep=0,outer sep=0,fill=tabred] (d2) at ( .15,-1*.12) {};
                \node[circle,minimum size=5px,inner sep=0,outer sep=0,fill=tabgre] (d3) at ( .15, 0*.12) {};
                \node[circle,minimum size=5px,inner sep=0,outer sep=0,fill=taborg] (d4) at ( .15, 1*.12) {};
                \node[circle,minimum size=5px,inner sep=0,outer sep=0,fill=tabblu] (d5) at ( .15, 2*.12) {};
                \node[circle,minimum size=5px,inner sep=0,outer sep=0,fill=tabgre] (m1) at (-.15,-1*.12) {};
                \node[circle,minimum size=5px,inner sep=0,outer sep=0,fill=taborg] (m2) at (-.15, 0*.12) {};
                \node[circle,minimum size=5px,inner sep=0,outer sep=0,fill=tabblu] (m3) at (-.15, 1*.12) {};
                \draw[line width=1pt] (m1) -- (d3);
                \draw[line width=1pt] (m1) -- (d5);
                \draw[line width=1pt] (m2) -- (d2);
                \draw[line width=1pt] (m2) -- (d4);
                \draw[line width=1pt] (m3) -- (d1);
            \end{scope}
        
            \node at (0.0, 2.2) {\includegraphics[width=.5in]{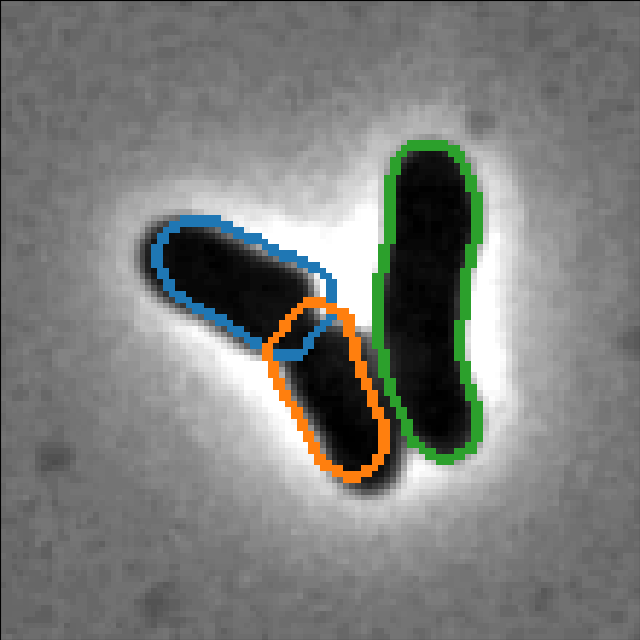}};
            \node at (1.0, 2.2) {\includegraphics[width=.5in]{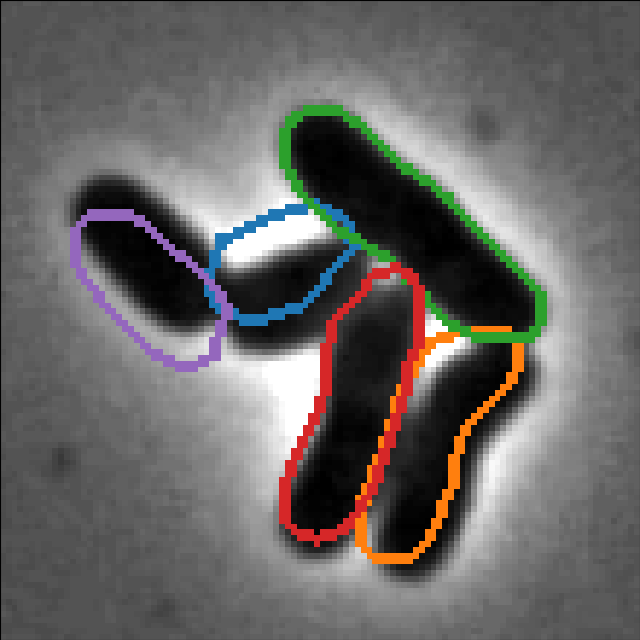}};
            \node[anchor=south west,white] at (0-.5, 2.4) {\scalebox{.8}{$t$}};
            \node[anchor=south west,white] at (1-.5, 2.4) {\scalebox{.8}{$t'$}};
            \node at (1.7, 2.2) {$\to$};

            \begin{scope}[cm={1.5, 0, 0, 1.5, (2.25, 2.2)}]
                \node[circle,minimum size=5px,inner sep=0,outer sep=0,fill=tabppl] (d1) at ( .15,-2*.12) {};
                \node[circle,minimum size=5px,inner sep=0,outer sep=0,fill=tabred] (d2) at ( .15,-1*.12) {};
                \node[circle,minimum size=5px,inner sep=0,outer sep=0,fill=tabgre] (d3) at ( .15, 0*.12) {};
                \node[circle,minimum size=5px,inner sep=0,outer sep=0,fill=taborg] (d4) at ( .15, 1*.12) {};
                \node[circle,minimum size=5px,inner sep=0,outer sep=0,fill=tabblu] (d5) at ( .15, 2*.12) {};
                \node[circle,minimum size=5px,inner sep=0,outer sep=0,fill=tabgre] (m1) at (-.15,-1*.12) {};
                \node[circle,minimum size=5px,inner sep=0,outer sep=0,fill=taborg] (m2) at (-.15, 0*.12) {};
                \node[circle,minimum size=5px,inner sep=0,outer sep=0,fill=tabblu] (m3) at (-.15, 1*.12) {};
                \draw[line width=1pt] (m1) -- (d3);
                \draw[line width=1pt] (m1) -- (d4);
                \draw[line width=1pt] (m2) -- (d2);
                \draw[line width=1pt] (m2) -- (d5);
                \draw[line width=1pt] (m3) -- (d1);
            \end{scope}

            \draw [very thick,decorate,decoration={calligraphic brace,amplitude=10pt}] (2.6, 2.75) -- (2.6, -.55);
            
            \node[align=center] at (3.4, 2.1) {\footnotesize Edge proba- \\[-1mm] \footnotesize bilities $P$, \\[-1mm] \footnotesize Eq.~\eqref{eq:feat-perturb}};
            \begin{scope}[cm={1.8, 0, 0, 1.8, (3.4, 1.1)}]
                \node[circle,minimum size=7px,inner sep=0,outer sep=0,fill=tabppl] (d1) at ( .15,-2*.12) {};
                \node[circle,minimum size=7px,inner sep=0,outer sep=0,fill=tabred] (d2) at ( .15,-1*.12) {};
                \node[circle,minimum size=7px,inner sep=0,outer sep=0,fill=tabgre] (d3) at ( .15, 0*.12) {};
                \node[circle,minimum size=7px,inner sep=0,outer sep=0,fill=taborg] (d4) at ( .15, 1*.12) {};
                \node[circle,minimum size=7px,inner sep=0,outer sep=0,fill=tabblu] (d5) at ( .15, 2*.12) {};
                \node[circle,minimum size=7px,inner sep=0,outer sep=0,fill=tabgre] (m1) at (-.15,-1*.12) {};
                \node[circle,minimum size=7px,inner sep=0,outer sep=0,fill=taborg] (m2) at (-.15, 0*.12) {};
                \node[circle,minimum size=7px,inner sep=0,outer sep=0,fill=tabblu] (m3) at (-.15, 1*.12) {};

                \draw[line width=1.5pt,black!10] (m3) -- (d5);
                \draw[line width=1.5pt,black!10] (m2) -- (d5);
                \draw[line width=1.5pt,black!10] (m1) -- (d5);
                
                \draw[line width=1.5pt,black!10] (m1) -- (d4);

                \draw[line width=1.5pt,black!45] (m2) -- (d4);

                \draw[line width=1.5pt] (m1) -- (d3);
                \draw[line width=1.5pt] (m2) -- (d2);
                \draw[line width=1.5pt] (m3) -- (d1);
            \end{scope}

        \end{tikzpicture}
    }
    \label{fig:feat-perturb}
    \caption{
        Schematic presentation of the feature perturbation approach. 
        As an example, we perturb segmentation masks by shifting them randomly,
        which defines a particular noise distribution $\noisedist$.
        For each perturbed pair of frames we compute the corresponding
        assignment solutions $\hat{A}$.
        The posterior edge probability is approximated as the
        frequency at which an edge appears in the perturbed solutions.
    }
    \vspace{-1mm}
\end{figure}

An alternative approach to create a set of $K$ candidate solutions, 
which mitigates the problem of sampling or enumerating
the set of valid assignments, is \emph{feature perturbation}.
Inspired by test-time augmentation \cite{wang_automatic_2019,moshkov_test-time_2020,wang_test-time_2022,kimura_understanding_2024}, 
we resample cellular features $\hat{\cell} \sim \noisedist(\hat{\cell} \cond \cell)$ from
some noise distribution $\noisedist$ conditional on the actual observations.
A simple example of a noise distribution for positional features is a Gaussian 
perturbation centered on the cell detections, \ie
$\hat{\cell} \sim \cell + \epsilon$ with $\epsilon \sim \normal(0, \gamma I).$
Given $K$ sets of perturbed features $\hat{\cells}_t^{(1)}, \ldots, \hat{\cells}_t^{(K)}$ 
and $\hat{\cells}_{t'}^{(1)}, \ldots, \hat{\cells}_{t'}^{(K)}$
for each frame $t, t'$,
\ie were the features of each cell have been resampled from the noise distribution,
we solve the corresponding assignments using the perturbed features
to obtain an ensemble of $K$ assignment solutions $\assignment_1, \ldots, \assignment_K$.
Similar to the approach from \Cref{sec:bayesian-assignment}, we use that
ensemble of solutions to construct a Monte Carlo estimator
\begin{align}
    \label{eq:feat-perturb}
    P_{ij}
        &\approx \frac{1}{K} \sum_{k=1}^K \ 
            \ind{(\cell_i, \cell_j') \in \hat{A}_k},
\end{align}
with 
$\hat{A}_k := \argmax_{\assignment \in \assignments} 
    \prob(\hat{\cells}_{t'}^{(k)} \cond \hat{\cells}_{t}^{(k)}, \assignment) $
to obtain a confidence score for any particular mother-daughter relationship.
We sketch this method in \Cref{fig:feat-perturb}.
Note that this approach requires the solution of $K$ assignment problems.
As a heuristic alternative, we also investigate performing the aggregation 
over the $K$ perturbed samples before solving the assignment, \ie we solve the
LAP for the mean cost function
\begin{align}
    \label{eq:perturbed-cost}
    \hat{w}(\cell_i, \cell_j') = \frac{1}{K} \sum_{(\hat{\cell}_i, \hat{\cell}_j') \in \hat{\cells}}
        w(\hat{\cell}_i, \hat{\cell}_j')
\end{align} 
only once.
Since for this approach we cannot count the frequency of given edges in the 
perturbed solutions as in (\ref{eq:feat-perturb}) before, we apply the 
classification-based approach introduced in \Cref{sec:classification} to get 
probabilistic confidence scores.

\subsubsection{Bayesian Neural Perturbations}
\label{sec:bayesian-feature-acquisition}

A key question in the feature perturbation approach is the specific
choice of noise distribution $\noisedist$
and its parameters, \eg the variance $\gamma$ in our previous
example of a Gaussian perturbation of the positions.
A particularly interesting approach to this question can be derived
from a probabilistic perspective at the feature extraction.
More specifically, if we have access to a stochastic feature extractor,
we may use its predictive distribution as our noise distribution $\noisedist$.
For this, we consider the feature extractor to be a neural network
$\nn : \img \to \R^d$ mapping raw images 
from the image space $\img$ to some latent space $\R^d$.
The neural network in this case may either be a segmentation model
\cite{ronneberger_u-net_2015, scherr_cell_2020, lee_mediar_2022, upschulte_contour_2022, upschulte_uncertainty-aware_2023}
or the neural network from the earlier discussed cost function in \Cref{eq:nn-cost}.

One conceptually attractive framework to obtain probabilistic predictions
from the neural feature extractor is that of \emph{Bayesian neural networks} 
\cite{mackay_probable_1995}.
Given some loss function $\ell_{\data}(\theta)$ 
over some dataset $\data$
with which the neural network is trained, 
we consider the posterior distribution
\begin{align}
    \label{eq:bnn-posterior}
    \pi(\theta \cond \data) \propto \likeli(\data \cond \theta) \cdot p(\theta)
\end{align}
of the neural network parameters $\theta$.
We do so by reinterpreting the loss as the negative log-likelihood 
$\ell_{\data}(\theta) \propto -\log \likeli(\data \cond \theta)$
and choosing a prior distribution $p(\theta)$ over neural network parameters.
Since the posterior distribution $\pi(\theta \cond \data)$ of neural networks 
is generally intractable in practice, one has to use an approximation instead.
Typical approximation techniques are mean-field variational inference 
\cite{graves_practical_2011, blundell_weight_2015},
Monte Carlo Dropout \cite{gal_dropout_2016} or 
Laplace approximations \cite{mackay_bayesian_1992, laplace_memoire_1774}.
Given the resulting approximate, but tractable surrogate posterior $q(\theta)$, 
the predictive distribution of the Bayesian neural network is approximated
by generating an ensemble $\{ f_{\theta_k} \}_{k=1}^K$ with $\theta_k \sim q(\theta)$ for $ k=1,\ldots,K \}$ 
of differently parametrized neural networks sampled from $q(\theta)$.
This ensemble

For feature perturbation in cell tracking, we use the ensemble to generate
a discrete distribution of features using the neural network ensemble. 
For the neural network-based cost function (\cf \Cref{eq:nn-cost}), the varying
parametrization of the neural networks results in feature perturbation in the 
latent space onto which the neural network encodes the incoming features. 

\newsavebox{\classificationfigure}
\savebox{\classificationfigure}{
    \centering
    \resizebox{1\columnwidth}{!}{
    \begin{tikzpicture}[x=.825in, y=.825in]
        \node[anchor=north] at (0.0, 1.7) {Mothers};
        \node[anchor=north] at (1.0, 1.7) {Daughters};
        
        \node at (0.0, 1.0) {\includegraphics[width=.75in]{t0.png}};
        \node at (1.0, 1.0) {\includegraphics[width=.75in]{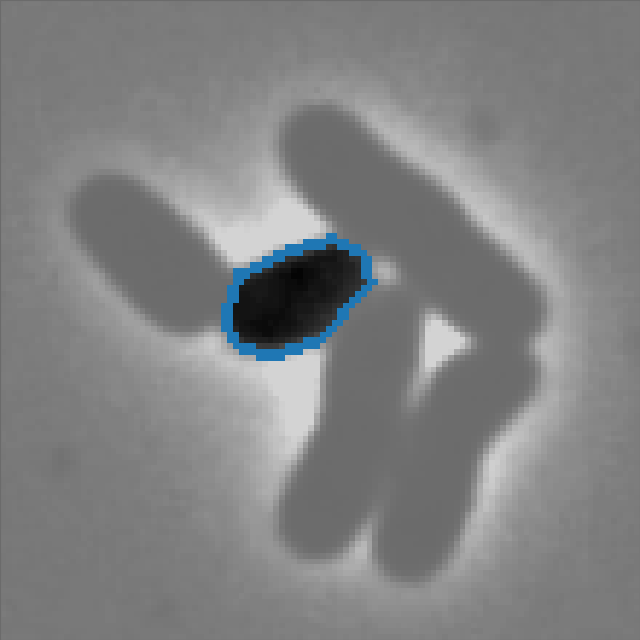}};
        \node[anchor=south west,white] at (0-.49, 1.25) {\scalebox{1}{$t$}};
        \node[anchor=south west,white] at (1-.49, 1.25) {\scalebox{1}{$t'$}};
        \draw [very thick,decorate,decoration={calligraphic brace,amplitude=5pt}] (1.5, 0.5) -- (-0.5, 0.5);
        \node at (0.5, 0.25) {$
                \begin{tikzpicture}
                    \node {};
                    \draw[line width=2pt,tabblu] plot coordinates {(0.0, 0) (0.3, 0)};
                \end{tikzpicture} \
                P_{\cdot|\begin{tikzpicture}\node[circle,minimum size=3.6px,inner sep=0,outer sep=0,fill=tabblu] {};\end{tikzpicture}}
                = (
                    P_{\begin{tikzpicture}\node[circle,minimum size=3.6px,inner sep=0,outer sep=0,fill=tabblu] {};\end{tikzpicture}|    
                        \begin{tikzpicture}\node[circle,minimum size=3.6px,inner sep=0,outer sep=0,fill=tabblu] {};\end{tikzpicture}},
                    P_{\begin{tikzpicture}\node[circle,minimum size=3.6px,inner sep=0,outer sep=0,fill=taborg] {};\end{tikzpicture}|
                        \begin{tikzpicture}\node[circle,minimum size=3.6px,inner sep=0,outer sep=0,fill=tabblu] {};\end{tikzpicture}},
                    P_{\begin{tikzpicture}\node[circle,minimum size=3.6px,inner sep=0,outer sep=0,fill=tabgre] {};\end{tikzpicture}|
                        \begin{tikzpicture}\node[circle,minimum size=3.6px,inner sep=0,outer sep=0,fill=tabblu] {};\end{tikzpicture}}
                )
            $};

        \def\x{0};

        \begin{scope}[cm={1, 0, 0, 1, (0, -1.5)}]

        \node at (\x+0.0, 1.0) {\includegraphics[width=.75in]{t0.png}};
        \node at (\x+1.0, 1.0) {\includegraphics[width=.75in]{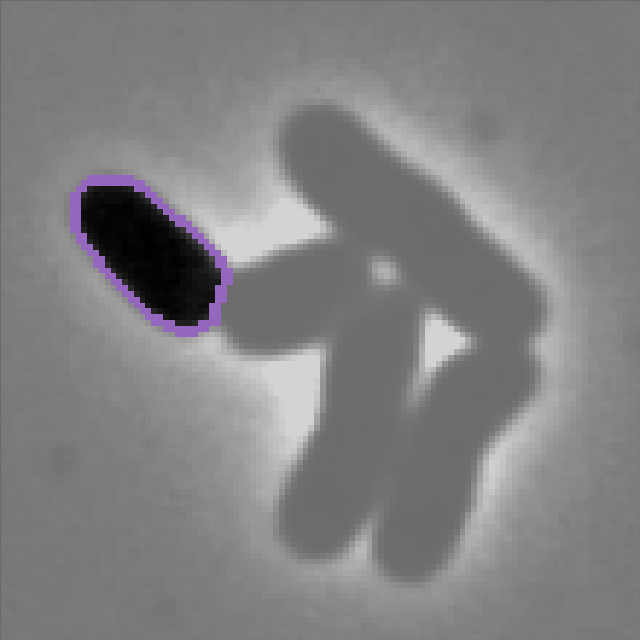}};
        \node[anchor=south west,white] at (\x+0-.49, 1.25) {\scalebox{1}{$t$}};
        \node[anchor=south west,white] at (\x+1-.49, 1.25) {\scalebox{1}{$t'$}};
        \draw [very thick,decorate,decoration={calligraphic brace,amplitude=5pt}] (\x+1.5, 0.5) -- (\x-0.5, 0.5);
        \node at (\x+0.5, 0.25) {$
                \begin{tikzpicture}
                    \node {};
                    \draw[line width=2pt,tabppl] plot coordinates {(0.0, 0) (0.3, 0)};
                \end{tikzpicture} \
                P_{\cdot|\begin{tikzpicture}\node[circle,minimum size=3.6px,inner sep=0,outer sep=0,fill=tabppl] {};\end{tikzpicture}}
                = (
                    P_{\begin{tikzpicture}\node[circle,minimum size=3.6px,inner sep=0,outer sep=0,fill=tabblu] {};\end{tikzpicture}|    
                        \begin{tikzpicture}\node[circle,minimum size=3.6px,inner sep=0,outer sep=0,fill=tabppl] {};\end{tikzpicture}},
                    P_{\begin{tikzpicture}\node[circle,minimum size=3.6px,inner sep=0,outer sep=0,fill=taborg] {};\end{tikzpicture}|
                        \begin{tikzpicture}\node[circle,minimum size=3.6px,inner sep=0,outer sep=0,fill=tabppl] {};\end{tikzpicture}},
                    P_{\begin{tikzpicture}\node[circle,minimum size=3.6px,inner sep=0,outer sep=0,fill=tabgre] {};\end{tikzpicture}|
                        \begin{tikzpicture}\node[circle,minimum size=3.6px,inner sep=0,outer sep=0,fill=tabppl] {};\end{tikzpicture}}
                )
            $};
        \end{scope}

        \node at (1.85, 0.075) {$\to$};

        \begin{scope}[cm={.35, 0, 0, 1., (2.2, -.5)}]
        \def\x{0};
        \def\y{0};

        \draw[->] (\x+0,\y+0) -- (\x+0.0, \y+1.5);
        \draw[->] (\x+0,\y+0) -- (\x+4, \y+0);
        
        \draw[line width=2pt,tabppl,fill=tabppl,fill opacity=.1] plot coordinates {
            (\x+0.5-.025, \y) 
            (\x+0.5-.025, \y+1.5-.2) 
            (\x+1.5-.025, \y+1.5-.2) 
            (\x+1.5-.025, \y+0.8-.2) 
            (\x+2.5-.025, \y+0.8-.2) 
            (\x+2.5-.025, \y+0.3-.2) 
            (\x+3.5-.025, \y+0.3-.2) 
            (\x+3.5-.025, \y) 
        };
        \draw[line width=2pt,tabblu,fill=tabblu,fill opacity=.1] plot coordinates {
            (\x+0.5+.025, \y) 
            (\x+0.5+.025, \y+1.3-.6) 
            (\x+1.5+.025, \y+1.3-.6) 
            (\x+1.5+.025, \y+1.4-.6) 
            (\x+2.5+.025, \y+1.4-.6) 
            (\x+2.5+.025, \y+1.0-.6) 
            (\x+3.5+.025, \y+1.0-.6) 
            (\x+3.5+.025, \y) 
        };

        \node[circle,minimum size=7px,inner sep=0,outer sep=0,fill=tabblu] at (\x+1, \y-.2) {};
        \node[circle,minimum size=7px,inner sep=0,outer sep=0,fill=taborg] at (\x+2, \y-.2) {};
        \node[circle,minimum size=7px,inner sep=0,outer sep=0,fill=tabgre] at (\x+3, \y-.2) {};
        \node at (\x+4, \y-.22) {$\cells_{t'}$};
        \end{scope}
    \end{tikzpicture}
    }
}
\subsection{Cell Tracking as Classification Task}
\label{sec:classification}

Given that sampling or enumeration of solutions to the LAP problem from \Cref{eq:ilp_obj}
is costly and can, thus, become 
infeasible,
we search for alternative approaches to perform uncertainty quantification in tracking.
In this section, we view the tracking problem as a classification problem, 
which recently has been leveraged to learn cost functions $w: \cells_t \times \cells_{t'} \to \R$
by means of neural networks~\cite{ben-haim_graph_2022,gallusser_trackastra_2024}.
We consider the classification perspective to leverage known techniques for uncertainty 
quantification and calibration for the cell tracking problem.

\begin{figure}
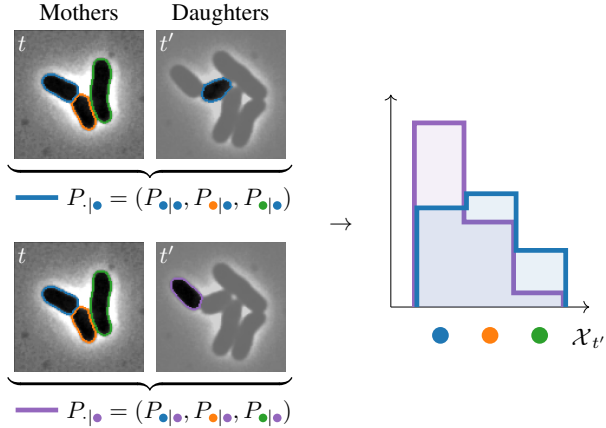

    \usebox{\classificationfigure}
    \caption{
        Two example DBMC class distributions. 
        The blue cell from frame $t'$ is located between the blue and orange cells 
        from frame $t$, thus assuming distance-based tracking this results in high 
        probability to be daughter of either of the two cells.
        In contrast, the purple cell is almost in the exact some spot as the blue
        cell is in frame $t$, resulting in high probability only for that mother.
    }
    \label{fig:dbmc}
\end{figure}

\subsubsection{Daughter-based Mother Classification}
\label{sec:mother-classification}

Cell tracking can be viewed as a multi-class classification
problem by considering the problem of choosing the right
mother for each individual daughter, 
which we refer to as \emph{daughter-based mother classification} (DBMC).
That is, for your favorite daughter $\cell_j' \in \cells_{t'}$, the \emph{classes} 
are just the mothers $\cells_{t}$ from the preceding frame.
The class -- or rather mother -- probabilities are obtained by performing 
daughter-
wise softmax normalization of the predicted costs
\begin{align}
    \label{eq:softmax}
    P_{i|j} &= 
    \prob((\cell_i, \cell_j') \in \assignment | \cell_j', \cells_t, \cells_{t'}) \\
    &=
            \frac{\exp{-w(\cell_i, \cell_j')}}{\sum_{\cell \in \cells_t} \exp{-w(\cell, \cell_j')}}.
\end{align}
From a generative perspective, this distribution could be used to sample
daughter-mother associations.
By construction, this approach guarantees that every daughter does indeed choose
only at most one mother, but it 
does not guarantee that every mother has at most two daughters,
as more than just two daughters might choose the same mother.

The DBMC perspective
enables straightforward setup of training pipelines for neural network and other parameters 
involved in the design of the cost function $w$,
as has been done in \cite{gallusser_trackastra_2024}.
Moreover, as we elaborate below,
we can interpret the softmax probabilities $P_{i|j}$ from
\Cref{eq:softmax} as confidence scores,
enabling their usage for uncertainty estimation.
In practice, using the \emph{parental softmax}
recently proposed by \citet{gallusser_trackastra_2024}
is
recommended 
to handle appearing cells, that \eg enter the image across its border.
We discuss the statistical assumptions introduced from parental softmax in \Cref{app:parental-softmax}.

Finally, the DBMC approach is also compatible to the Bayesian approaches from
\Cref{sec:bayesian-inference} by simply column-normalizing the edge 
probabilities $P_{ij}$ from 
\Cref{eq:importance-weighted} and \Cref{eq:feat-perturb}, 
\ie
\begin{align}
    \label{eq:column-normlized}
    P_{i|j} = \frac{P_{ij}}{\sum_{k} P_{kj}}.
\end{align}

\subsubsection{Calibration}
\label{sec:calibration}

A typical measure to assess the reliability of uncertainty-aware predictions is \emph{calibration}.
A classifier producing a 
prediction $y$ with attached confidence score $s \in [0, 1]$
is said to be \emph{perfectly calibrated}, if the confidence score matches 
the empirical probability of the result to be correct \cite{guo_calibration_2017}, 
\ie the probability to have the correct result $y^*$ should be exactly $s$, \ie
\begin{align}
    \label{eq:calibration}
    \prob(y = y^* \cond s) 
        = s.
\end{align}
The lack of calibration is quantified by considering the 
\emph{expected calibration error} (ECE)~\cite{guo_calibration_2017}
\begin{align}
    \mathbb{E}\left[|\prob(y = y^* \cond s) - s|\right].
\end{align}
In practice, calibration has to be assessed by binning predictions based on the
confidence score as presented in \cite{guo_calibration_2017}.

In order to consider calibration in cell tracking, we leverage the
introduced DBMC perspective.
That is, given a predicted daughter-mother pair $(\cell_i, \cell_j') \in \assignment$,
the confidence 
on this decision is given as the mother probability $P_{i|j}$.
Note that this is readily applicable not only to the DBMC approach based on the 
cost matrix as in \Cref{eq:softmax}, but also to the Bayesian treatments based on
the column-normalization from \Cref{eq:column-normlized}.
However, since the Bayesian treatments generate distributions of solutions, rather than
a single solution, it may remain ambiguous what we consider a 'prediction'.
A natural candidate in this setting is to consider the MAP estimate $\optassignment$
from \Cref{eq:optimize}.

\subsubsection{Temperature Scaling}
\label{sec:temp-scaling}

A widely used post-hoc technique to improve calibration
of classifiers is \emph{temperature scaling} \cite{guo_calibration_2017},
which introduces a temperature parameter $\tempparam > 0$, \st
\begin{align}
    \label{eq:temp-scaling}
    P_{i|j\!\tempparam} = \frac{P_{ij}^{\tempparam}}{\sum_k P_{kj}^{\tempparam}}.
\end{align}
The exponent $\tempparam$ is 
computed by optimizing the cross entropy loss between predicted and true class distribution 
\cite{guo_calibration_2017}.

It is worth noting that temperature scaling for the distance-based cost function $w_{_{\text{L2}}}$
from \Cref{eq:cost} effectively scales the variance of the Brownian motion~(\cf \Cref{app:cost-functions}).
For other cost functions that are based on a $p$-norm distance between possibly latent cell features,
as \eg in \Cref{eq:nn-cost}, it also scales the variance of the implicitly assumed
generalized normal likelihood function $\prob(x_j' \cond x_i)$
of said latent features (\cf \Cref{app:cost-functions}).

\subsubsection{Quantifying Tracking Uncertainty}
\label{sec:quantifying}

In the DBMC approach, we use the conditional probability $P_{i|j}$ as the probability
for $\cell_i$ being the mother of a given cell $\cell_j'$ among all candidates $\cells_t$.
Given an assignment solution $\assignment$, then the probability for any edge -- 
a daughter-mother pair -- to be correct is just the probability $P_{i|j}$ and thus a low
probability is linked to high uncertainty.

An alternative approach to quantify the uncertainty within 
DBMC 
is to
consider the \emph{daughter-wise entropy}
\begin{align}
    \mathcal{H}_j = - \sum_i P_{i|j} \log P_{i|j}
\end{align}
of the conditional distribution $P_{i|j}$.
High entropy means that the distribution is closer to uniformity and, thus, the decision
is ambiguous \cite{hullermeier_aleatoric_2021}.
However, since entropy is not a probability, it does not fit into the earlier discussed
calibration framework.
Nevertheless, entropy can be used as a decision criterion to distinguish possibly erroneous
decisions from correct ones.

\section{Related Work}
\label{sec:related-work}

This work is not the first to address cell tracking in terms of 
probabilistic methods.
Already the seminal work by \citet{crocker_methods_1996} phrased tracking
as a maximum likelihood problem assuming Brownian particle motion.
In this work, we challenge this viewpoint, arguing that the commonly used
methods are rather to be seen as inferring a Bayesian MAP estimate, 
which not only nicely accommodates biological feasibility within the 
prior distribution, but also lays the foundation to argue about
uncertainty in the tracking solution using the arising posterior distribution.

Work by \citet{theorell_when_2019} and \citet{kaiser_cell_2024} is close to ours 
in that they also model tracking uncertainty by keeping track of multiple 
possible tracking hypotheses.
Nevertheless, our work differs from theirs substantially in that we provide
a very general framework for probabilistic tracking under a Markov assumption,
\ie independence between the tracking solutions from frame $t$ to $t+1$ and 
those from $t+1$ to $t+2$.
In contrast, \citet{theorell_when_2019} and \citet{kaiser_cell_2024} specifically 
overcome the Markov assumption by informing the cost function for frame 
$t$ to $t+1$ based on the set of tracking hypotheses for all frames up to time $t$.

Concurrently to our work, \citet{betjes_cell_2024} presented a method
for \emph{``cell tracking with accurate error prediction''} where link
likelihoods predicted by a neural network are calibrated by enumerating
alternative assignment solutions within a small neighborhood of adjacent
cells and applying temperature scaling.
Our work generalizes their approach to a wider family of 
tracking algorithms based on our plug-and-play-like framework.
Moreover, we evaluate uncertainty estimation based on the DBMC approach,
whereas \citet{betjes_cell_2024} do so by treating tracking as a binary
edge classification problem.
Finally, their local enumeration algorithm may be a promising 
alternative to the global enumeration method used within this work in
\Cref{sec:bayesian-assignment}.

Approaching cell tracking as a multi-class classification problem 
has been considered before 
\cite{gallusser_trackastra_2024}.
We build upon that idea to quantify tracking uncertainty and calibration, 
which to the best of our knowledge has not yet been considered.
In addition, we give some statistical considerations on the parental softmax
introduced in \cite{gallusser_trackastra_2024}.

\section{Experimental Evaluation \& Discussion}
\label{sec:experiments}

Our methods for complementing cell tracking with uncertainty quantification work 
as a framework, which accommodates existing linear assignment-based 
TbD algorithms by simply specifying the corresponding
weight function $w$.
In the following experimental evaluation, we consider activity-based 
\cite{ruzaeva_cell_2022}, distance-based  \cite{crocker_methods_1996, fukai_laptrack_2023}, 
overlap-based  \cite{fukai_laptrack_2023}, and Transformer-based tracking \cite{gallusser_trackastra_2024},
each defining a different cost function $w$ (\cf~\Cref{app:cost-functions}).
We test the different methods introduced in \Cref{sec:framework} and summarized
in \Cref{tab:methods}.

\begin{figure*}
    \centering
    \begin{tikzpicture}
    \node at (0, 1.4) {\includegraphics[width=\columnwidth]{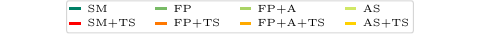}};
    \end{tikzpicture}
    \vspace{-7mm}
    
    \begin{subcaptiongroup}
        \subcaptionlistentry{}
        \label{fig:ece}
        \begin{tikzpicture}
            \node at (-.52\columnwidth, 0) {\includegraphics[width=\columnwidth]{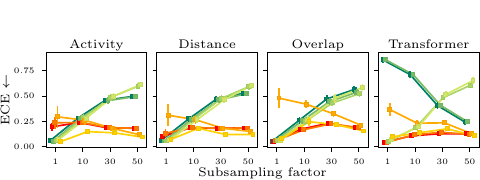}};
            \node at (-\columnwidth, .8) {\captiontext*{}};
        \end{tikzpicture}~%
        \subcaptionlistentry{}~%
        \label{fig:temp}~%
        \hspace{.1\columnsep}~%
        \begin{tikzpicture}
            \node at (-.52\columnwidth, 0) {\includegraphics[width=\columnwidth]{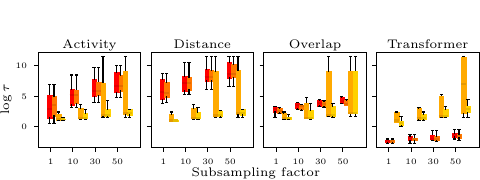}};
            \node at (-\columnwidth, .8) {\captiontext*{}};
        \end{tikzpicture}
    \end{subcaptiongroup}
    \vspace{-4mm}
    \caption{
        (a) Mean ECE and standard error thereof across datasets at decreasing
            temporal resolution.
        (b) Box plots showing the distributions of optimized temperatures at 
            decreasing temporal resolution. For non-tempered
            methods, the temperature is $\log\tau = 0$, implicitly.
    }
    
    \label{fig:calibration}
\end{figure*}

\begin{table}[]
    \centering
    \caption{
        Overview of our methods for probabilistic cell tracking: 
        Softmax'ing (SM), 
        Feature Perturbation (FP),
        Feature Perturbation with Assignment (FP+A) and
        Assignment Sampling (AS).
        For all methods, we also test using temperature scaling, denoted by a trailing +TS in the abbreviation.
        To refer to temperature scaling in general, we use TS \protect\temp. 
    }
    \resizebox{\columnwidth}{!}{
    \begin{tabular}{r|c|c}
        Method & $P_{ij}$ & \cf Eq. \\
        \hline\hline && \\[-3.5mm]
        SM      \SMm      & $\exp{-w(\cell_i, \cell_j')}$                                                & \eqref{eq:softmax} \\[2mm]
        FP      \FPm      & $\exp{-\hat{w}(\cell_i, \cell_j')}$                                          & \eqref{eq:perturbed-cost} \\[2mm]
        FP+A    \FPAm     & $K^{-1} \sum_{k=1}^K \ind{(\cell_i, \cell_j') \in \hat{\assignment}_k}$      & \eqref{eq:feat-perturb} \\[2mm]
        AS      \ASm      & $\sum_{k=1}^{K} \ind{(\cell_i, \cell_j') \in \assignment_k} \cdot p_k$       & \eqref{eq:importance-weighted} \\[2mm]
        \hline && \\[-3.5mm]
        SM+TS   \SMTSm    & $\exp{-\!\tempparam\!w(\cell_i, \cell_j')}$                                                                  & \eqref{eq:softmax}, \eqref{eq:temp-scaling} \\[2mm]
        FP+TS   \FPTSm    & $\exp{-\!\tempparam\!\hat{w}(\cell_i, \cell_j')}$                                                            & \eqref{eq:perturbed-cost}, \eqref{eq:temp-scaling} \\[1mm]
        FP+A+TS \FPATSm   & $\left( K^{-1} \sum_{k=1}^K \ind{(\cell_i, \cell_j') \in \hat{\assignment}_k} \right)^{\!\tempparam}$       & \eqref{eq:feat-perturb}, \eqref{eq:temp-scaling} \\[1mm]
        AS+TS   \ASTSm    & $\left( \sum_{k=1}^{K} \ind{(\cell_i, \cell_j') \in \assignment_k} \cdot p_k \right)^{\!\tempparam}$        & \eqref{eq:importance-weighted}, \eqref{eq:temp-scaling} 
    \end{tabular}
    }
    \label{tab:methods}
\end{table}

For the feature perturbation-based methods, we use Gaussian perturbations of the cell centers with 
variance $\gamma = 0.1$ for the distance- and activity-based algorithms, 
a randomized inflation or deflation of the cell area by thickening or thinning the cell borders
for the overlap-based algorithm, and a Bayesian version of the Transformer-based algorithm
using MC Dropout \cite{gal_dropout_2016}.
For the latter, we test dropout rates 0.5, 0.25, and 0.125.
For the assignment sampling and feature perturbation strategy we take $K=10$ samples.

The datasets under consideration are 2D datasets from the 
Cell Tracking Challenge \cite{maska_benchmark_2014, maska_cell_2023}
and a publicly available dataset with five fully annotated time-lapse sequences of microbial 
colonies \cite{seiffarth_data_2022}.
We used the ground truth segmentations as detections for the tracking, thus assuming perfect
segmentation, isolating the effect of uncertainty intrinsic to the tracking problem.
Since each dataset consisted of at least two sequences, we split all datasets such that one
sequence was used for temperature scaling (where applicable) and the remaining sequences
were used to evaluate our methods.
Moreover, to simulate low temporal resolution, we also subsample the sequences by only considering
every 10th, 30th, and 50th image from a time-lapse sequence.

\subsection{Calibration}
\label{sec:calibration-results}

We focus our evaluation on the question of whether our approaches to uncertainty-aware tracking
are capable of reporting 'useful' uncertainties.
Calibration (\cf \Cref{sec:calibration}) tests for a one-to-one correspondence between
estimated confidence and empirically achieved average accuracy.
Thus, for a well-calibrated predictor, the estimated confidence is a predictor of the chance
for the result to be correct.
In \Cref{fig:ece}, we depict the expected calibration error (ECE) averaged across the
used datasets of our methods for various tracking algorithms and for varying temporal resolution.
Most notably, we observe that temperature scaling (TS \temp) almost always improves the ECE
of any of our uncertainty-aware tracking methods,
especially at the lowest temporal resolutions tested.
The \ASTS method yields particular good calibration across all tested algorithm and temporal
resolutions.

For all non-tempered methods, except for \SM and \FP using the Transformer-based tracking, 
we do observe declines in calibration as the temporal resolution
decreases. 
A possible explanation for the described decline in calibration,
visible by an increase in ECE, may be our earlier consideration
that the temperature represents the average squared displacement across time, as we discussed
for the distance-based tracking (\cf \Cref{sec:temp-scaling} \& \Cref{app:cost-functions}).
Hence, as the temporal resolution decreases, the average squared displacement between images 
increases and thus the implicitly assumed temperature of $\tau=1$ does not match to the increasing
squared displacement.

We verify this explanation by comparing the optimized temperature to the mean squared
displacements observed at given temporal resolution in \Cref{fig:temp-displace},
which, indeed, suggests a positive correlation between the two.
This observation further implies that improved calibration is achieved if the 
Brownian motion model underlying the distance-based tracking also models
the motion's variance correctly, which is usually neglected.
Moreover, for the other tracking algorithms, where the temperature parameter does not admit
a physical interpretation, we depict the distribution of optimized temperatures across datasets
against the decrease in temporal resolution in \Cref{fig:temp}.
As expected, the temperature raises as temporal resolution decreases, suggesting that less confident
predictions improve calibration in the low resolution setting.
This observation is in line with our intuitive expectations, 
as even humans struggle with correctly tracking cells when temporal resolution is low,
due to the possibly large changes inbetween frames.

Interestingly, the increase in temperature is least notable for the \SM and \FP methods
when using Transformer-based tracking.
The same setups are the only non-tempered ones to achieve improved calibration as the 
temporal resolution declines (\cf \Cref{fig:ece}).
A possible explanation for this might be 
that the Transformer-based approach
computes its cost matrices for frame-to-frame associations as the off-diagonal elements
of a multi-frame block-wise cost matrix (\cf \Cref{app:cost-functions}) of encoded
features.
Those features have been encoded via an attention mechanism and, thus, already interacted
with each other across multiple frames.
This allows the Transformer to include multi-frame
information into its feature encoding and finally the cost matrix, thus, possibly 
calibrating the costs already based on the multiple, available frames.

\begin{figure}
    \centering
    \includegraphics[width=\columnwidth]{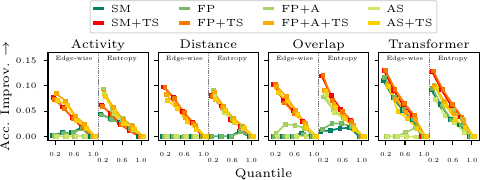}
    \caption{
        Mean accuracy improvement across tested datasets and varying
        temporal resolution
        (higher is better) using either the 
        \emph{edge-wise} probability or the daughter-wise \emph{entropy} 
        as a criterion to sparsify the prediction,
        \ie omit predictions, which exceed a certain sparsification theshold 
        computed as quantiles over the sparsification criterion.
    }
    \label{fig:sparsification}
\end{figure}

\subsection{Sparsification}
\label{sec:sparsification}

Another perspective on the 'usefulness' of uncertainty estimates is given by 
\emph{sparsification plots}. 
Intuitively speaking, useful uncertainties are negatively correlated with the performance of a model, 
that is, high uncertainty corresponds to erroneous decisions, 
whereas low uncertainty corresponds to correct ones.
Sparsification plots analyze this correlation by gradually removing the most uncertain decisions, 
while observing the performance of the remaining more certain ones, which should ideally increase.
This type of analysis has particular relevance for practical applications of uncertainty estimation,
as it allows to estimate prediction improvements and required effort within a human-in-the-loop 
approach, where the most uncertain predictions are corrected manually by an expert.

In \Cref{fig:sparsification} we present such sparsification plots, where we show the average 
improvement in accuracy across different subsampling factors for our various uncertainty 
estimation methods and tracking algorithms.
We refer to \Cref{fig:sparsification-detailed} for a more fine-grained presentation of 
the sparsification results.
Moreover, we measure uncertainty using either edge-wise probabilities
or the daughter-wise entropy as discussed in 
\Cref{sec:quantifying}.

We note that in comparison to ECE (\cf \Cref{fig:ece}), TS~\temp is less
critical to achieve accuracy improvement, especially when using the daughter-wise entropy 
for uncertainty estimation.
In particular, the \FPA and \AS methods exhibit only slight increases in accuracy improvement 
when using TS (\cf \FPATS \& \ASTS) on top.
However, for the edge-wise uncertainty estimation, TS is crucial for both methods and across
all tested tracking algorithms in order to gain any accuracy improvements at all.
For the \SM and \FP methods on the other hand, TS seems to be only crucial for the overlap-based
tracking. 
In most other cases, TS does still yield accuracy improvements, but not as distinct.
This seems to be mostly independent of the uncertainty measure in use.

In summary, entropy-based uncertainty estimation is less reliant on TS \temp,
and, thus, on annotated training data.
A possible explanation for this might be that the edge-wise uncertainty estimation 
before TS does not incorporate any information on the class probabilities for the alternative
mothers at choice except of a normalization factor, 
whereas computing the entropy across those class probabilities does.
Nevertheless, TS \temp does still mostly yield improved sparsification performance and
should be, thus, used, if annotated training data is available.

\section{Conclusion}
\label{sec:conclusion}

In this work, we presented methods to quantify uncertainty in cell tracking
caused by the ambiguity arising from low temporal resolution and high similarity
of the objects which are to be tracked.
To this end, we leveraged the two perspectives -- as a Bayesian inference and
as a multi-class classification problem -- on cell tracking, which motivate the presented methods explored in this work.
Based on the classification perspective, we presented approaches to
quantify the reliability of our uncertainty-aware tracking methods in terms of
calibration and sparsification. 

Our experimental results indicate that tracking algorithms are not \emph{per se}
well-calibrated, but calibration is achievable using temperature scaling if 
annotated tracking data is available.
Nevertheless, we notice that using our proposed daughter-wise entropy approach for quantifying
uncertainty, even the vanilla tracking algorithms can produce 'useful' uncertainties,
\ie uncertainties that positively correlate with model performance. 
However, in particular our approximate Bayesian methods \FPA \& \AS
show improved performance in terms of sparsification and are less reliant
on temperature scaling and, thus, annotated tracking data,
though this comes at increased computational costs (\cf \Cref{fig:runtime}).

\subsection{Future Work}
\label{sec:future-work}

Although our experimental results indicate that our methods do already 
provide useful uncertainty estimation, we hope to further inspire the development
of uncertainty-aware tracking algorithms and incorporation of quantitative evaluation
on the usefulness of the estimated uncertainties.

Moreover, tuning hyperparameters of our methods, such as the noise distribution in the
feature perturbation approaches, which were fixed within this work,
motivate the development of automated hyperparameter tuning
pipelines for our uncertainty-aware tracking methods.

\section*{Acknowledgements}
RDP is funded by the Helmholtz School for Data Science in Life, Earth, and Energy (HDS-LEE). 
DR’s research is funded by the Deutsche Forschungsgemeinschaft 
(DFG, German Research Foundation) -- 548823575.
This work was
supported by the President’s Initiative and Networking Funds of
the Helmholtz Association of German Research Centres [EMSIG ZT-I-PF-04-044]. 
We also thank Ben Gallusser for his support and input.

{
    \small
    \bibliographystyle{ieeenat_fullname}
    \bibliography{main, references}
}

\newpage
\appendix
\onecolumn

{
    \centering
    \textbf{\Large 
        Supplemental Material:\\ 
        How To Make Your Cell Tracker Say: "I dunno!"
        \vspace*{\baselineskip}\\
    }
}

\section{Linear Assignment Formulation}
\label[app]{app:ilp}

Given two consecutive frames $\cells_t$ and $\cells_{t'}$, there are three possible associations 
for the cells in those two frames: A daughter cell $\cell' \in \cells_{t'}$ is linked to a mother
cell $\cell \in \cells_t$, a daughter cell remains without a mother or a mother cell remains 
without a daughter. 
The latter two cases are relevant \eg if cells enter or leave the region of interest from across 
the frame border.
Technically, we we can consider the cases, were mother or daughter cells are not associated to any
other cell as an association to some fallback class $\perp$ and denote the likelihood of a single
cell appearing or disappearing as $\prob(\cell_j' \cond \perp)$ and $\prob(\perp\cond\cell_i)$,
respectively.
Given those probabilities,
we denote the joint likelihood of observing detections $\cells_{t'}$ given $\cells_t$ and some
assignment solution $\assignment$ as
\begin{align}
    \label{eq:joint-likelihood}
        \prob(\cells_{t'} \cond \cells_t, \assignment) =
        \prod_{(\cell_i, \cell_j') \in \assignment} \prob(\cell_j' \cond \cell_i)
        \prod_{\cell_i \notin \assignment} \prob(\perp \cond \cell_i)
        \prod_{\cell_j' \notin \assignment} \prob(\cell_j' \cond \perp),
\end{align}
where any cell not included in the assignment $\assignment$ is either an appearing daughter
or disappearing mother.
Defining $w_a := \log \prob(\cell_j' \cond \perp)$, $w_d := \log \prob(\perp \cond \cell_i)$
and $m, n$ as the number of appearing and disappearing cells respectively, we obtain the
linear assignment formulation from \Cref{eq:ilp_obj} by taking the logarithm of our 
likelihood from \Cref{eq:joint-likelihood}.

\section{Tracking Cost Functions}
\label[app]{app:cost-functions}

The tracking algorithms used in this work are mainly distinguished by the cost function $w$, 
which they define.
Here, we give the remaining two cost functions, which were not introduced in the main paper,
as well as some details on the Transformer-based cost function and elaborate the connections 
between distance- and activity-based tracking and temperature scaling.

\begin{itemize}
    \item
In distance-based tracking, 
the cellular features are their positions in the image and the cost function
is computed as
\begin{align}
    w_{_\text{L2}}(\cell_i, \cell_j') = \frac{\lambda}{2}\|\cell_i - \cell_j'\|_2^2,
\end{align}
which is equivalent to assuming Brownian motion with variance $\lambda^{-1}$, \ie
$\cell_j' \sim \normal(\cell_i, \lambda^{-1} I)$.

    \item
The activity-based tracking is similar to the distance-based tracking, 
but introduces an additional activity value $\alpha_i$
for each mother cell $\cell_i$, which scales the variance of the Gaussian likelihood 
$\cell_j' \sim \normal(\cell_i, \alpha_i \lambda^{-1} I)$
\begin{align}
    w_{_\text{AC}}(\cell_i, \cell_j') = \frac{\lambda}{2 \alpha_i}\|\cell_i - \cell_j'\|_2^2.
\end{align}
The activity value is computed from the raw image values within the segmentation mask of $\cell_i$.
For details, please refer to the original publication \cite{ruzaeva_cell_2022}.

    \item
In the overlap-based tracking, the cellular features are their segmentation masks and the cost function
is computed as the negative number of pixels in the overlap, \ie
\begin{align}
    w_{_\text{OL}}(\cell_i, \cell_j') = -|\cell_i \cap \cell_j'|
\end{align}

    \item
The Transformer-based tracking algorithm \emph{Trackastra} \cite{gallusser_trackastra_2024} defines a frame-to-frame
cost function, but incorporates detections from multiple frames in a sliding window fashion.
Given a sliding window of size $\Delta \in \mathbb{N}$, \emph{Trackastra} encodes \emph{shallow} input features
$\cell_i^{(t+\delta)} \in \cells_{t+\delta}$ with $\delta=0, \ldots, \Delta$
using two different functions $f_{\theta}, g_{\theta}$ implement as Transformer neural networks.
Given a set of cellular features $\cells$, $f_{\theta}$ and $g_{\theta}$ map 
those features to latent spaces $\mathcal{Y}=f_{\theta}(\cells)$ and $\mathcal{Z}=g_{\theta}(\cells)$,
such that $\mathcal{Y}_i, \mathcal{Z}_i$ are the latent representations corresponding to a particular input feature $\cells_i \in \cells$.
Shallow features used here are position, shape descriptors and image intensities.
Frame-to-frame costs are then computed as 
\begin{align*}
    w(\cell_i^{(t')}, \cell_j^{(t'+1)}) = \| \mathcal{Y}_i^{(t')} - \mathcal{Z}_j^{(t'+1)} \|_2^2,
\end{align*}
where $\mathcal{Y}_i^{(t')}, \mathcal{Z}_j^{(t')}$ are the latent representations of 
$\cell_i^{(t)}, \cell_j^{(t+1)}$ respectively and $\mathcal{Y}, \mathcal{Z}$ were computed on
the union of all cellular features $\cells=\bigcup_{\delta=0}^\Delta \cells_{t+\delta}$ from the sliding window.

\end{itemize}

\bigskip\noindent
To show that temperature scaling effectively scales the variance of the Brownian motion assumed
by the distance- and activity-based tracking algorithm, we consider \Cref{eq:temp-scaling},
plug in \Cref{eq:softmax} and absorb the temperature into a scaled cost function:
\begin{align}
    P_{i|j\!\tempparam} 
        &= \frac{P_{ij}^{\tempparam}}{\sum_k P_{kj}^{\tempparam}} \\
        &= \frac{\exp{-\tempparam\!w(\cell_i, \cell_j')}}{\sum_{\cell_k \in \cells_t} \exp{-\tempparam\!w(\cell_k, \cell_j')}} \\
        &= \frac{\exp{-{\color{magenta}\tilde{w}}(\cell_i, \cell_j')}}
            {\sum_{\cell_k \in \cells_t} \exp{-{\color{magenta}\tilde{w}}(\cell_k, \cell_j')}},
\end{align}
where we define ${\color{magenta}\tilde{w}}(\cell_i, \cell_j') := \tempparam\!w(\cell_i, \cell_k')$.
Now setting $w = w_{_\text{L2}}$, we see that 
${\color{magenta}\tilde{w}}(\cell_i, \cell_j') = \frac{\tempparam\!\lambda}{2}\|\cell_i - \cell_j'\|_2^2$,
which is equivalent to assuming $\cell_j' \sim \normal(\cell_i, (\tempparam\!\lambda)^{-1}I)$.
Similarly, for
$w = w_{_\text{AC}}$, we get that
${\color{magenta}\tilde{w}}(\cell_i, \cell_j') = \frac{\tempparam\!\lambda}{2\alpha_i}\|\cell_i - \cell_j'\|_2^2$,
which is equivalent to assuming $\cell_j' \sim \normal(\cell_i, \alpha_i(\tempparam\!\lambda)^{-1}I)$.
Finally, if we set $w = \| f_{\theta}(\cell_i) - f_{\theta}(\cell_j) \|_p^p$, where $\| \cdot \|_p$ is the $p$-norm
and $f_{\theta}$ some function projecting our cell features into some latent space,
then this is equivalent to assuming $f_{\theta}(\cell_j') \sim \mathcal{GN}(f_{\theta}(\cell_i), \tempparam^{-1}I, p)$,
where $\mathcal{GN}(\mu, \alpha, \beta)$ is the generalized normal distribution with location $\mu$, scale $\alpha$
and shape $\beta$.

\section{Parental Softmax}
\label[app]{app:parental-softmax}

A practical issue that might arise from the
DBMC approach is caused by appearing cells, that \eg cross the image border.
In this case, the true probability for any detection $\cell_i$ from frame $t$ to 
be the mother of the appearing cell $\cell_j'$  should be zero. 
However, the plain softmax distribution from \Cref{eq:softmax} cannot capture
this case, as the denominator would become 0, if the tracking algorithm were to
correctly yield edge-wise probabilities $\exp{w(\cell_i, \cell_j')} = 0$.
Thus, \citet{gallusser_trackastra_2024} introduce the \emph{parental softmax}
\begin{align}
    \label{eq:parental-softmax}
    \prob((\cell_i, \cell_j') \in \assignment | \cell_j', \cells_t, \cells_{t'}) 
        = \frac{\exp{w(\cell_i, \cell_j')}}
             { 1+ \sum_{\cell \in \cells_t} \exp{w(\cell, \cell_j')}}.
\end{align}
by adding a constant to the denominator.
From a statistical perspective, this might seem like an issue, as the parental softmax
distribution does not sum up to 1 anymore.
In fact, however, this constant summand represents an unnormalized probability
for the implicit class $\perp$ representing the lack of an adequate mother.
The normalized probability of this choice can be computed as
\begin{align}
    \prob((\perp, \cell_j') \in \assignment | \cell_j', \cells_t, \cells_{t'}) 
    = 1 - \sum_{\cell \in \cells_t} \prob((\cell, \cell_j') \in \assignment | \cell_j', \cells_t, \cells_{t'}).
\end{align}

\section{Bayesian Tracking Transformer}
\label[app]{app:dropout}

\begin{figure}
    \centering
    \begin{minipage}[t][][b]{.26\textwidth}
        \centering
        \vspace{9.5mm}
        \includegraphics[width=\textwidth]{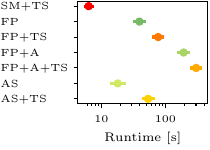}
        \caption{
            Mean runtime overhead and standard error thereof 
            compared to the SM method, which has virtually the
            same costs as vanilla uncertainty-unaware tracking.
        }
        \label{fig:runtime}
    \end{minipage}~\hspace{.05\textwidth}~
    \begin{minipage}[t][][b]{.36\textwidth}
        \centering
        \includegraphics[width=\textwidth]{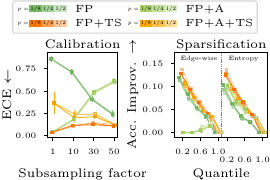}
        \caption{
            Calibration and sparsification as discussed in 
            \Cref{sec:calibration-results} and \ref{sec:sparsification}
            shown for the Transformer-based tracking using 
            Monte Carlo dropout at varying dropout rates.
        }
        \label{fig:dropout}
    \end{minipage}~\hspace{.05\textwidth}~
    \begin{minipage}[t][][b]{.26\textwidth}
        \centering
        \vspace{3mm}
        \includegraphics[width=\textwidth]{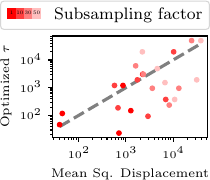}
        \caption{
            Optimized temperature $\tempparam$ of our \protect\SMTS method 
            plotted against the empirical
            mean squared displacement of cells based on the ground truth tracking
            for each.
            The dashed line depicts 1-to-1 correspondence.
        }
        \label{fig:temp-displace}
    \end{minipage}
\end{figure}

For the Bayesian Transformer-based tracking we trained \textit{Trackastra} \cite{gallusser_trackastra_2024}
using MC Dropout \cite{gal_dropout_2016} at varying dropout probabilities $p \in \{0.5, 0.25, 0.125\}$.
In \Cref{fig:dropout}, we present the ECE and sparsification as before in
\Cref{sec:calibration-results} \& \ref{sec:sparsification}, but across the tested dropout rates.

Our results indicate no clear trend, favoring neither lower nor higher dropout probabilities.
For the ECE, especially \FP and \FPA show only slight variance in performance depending on 
the dropout probability.
For sparsification, some variance is noticeable, though only the \FP method using entropy-based
uncertainty estimation indicates a trend favoring higher dropout probabilities.
For the temperature scaled versions (\FPTS \& \FPATS) ECE seems to be improved by a lower dropout 
probability, however an inverted trend is visible for sparsification, where high dropout probabilities
achieve higher accuracy improvements.

Given the recent advances in Bayesian deep learning, applying more sophisticated techniques like
Laplace approximations \cite{mackay_probable_1995, ritter_scalable_2022} or variational inference 
\cite{shen_variational_2024} might further improve the performance of our Bayesian neural perturbation 
method.

\begin{figure*}
    \centering
    \begin{tikzpicture}
    \node at (0, 0) {\includegraphics[width=.7\columnwidth]{legend.pdf}};
    \end{tikzpicture}
    
    \includegraphics[width=.8\textwidth]{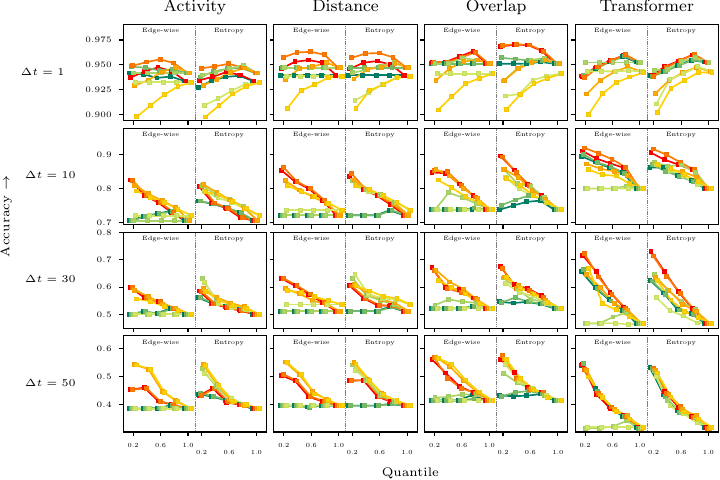}
    \caption{
        Mean accuracy of tracking predictions per temporal resolution, 
        sparsified at varying thresholds, 
        which were computed as quantiles over the respective
        sparsification criterion, \ie either the \emph{edge-wise} probability
        or the daughter-wise \emph{entropy}.
    }
    \label{fig:sparsification-detailed}
\end{figure*}

\begin{figure*}
    \centering

    \begin{tikzpicture}
    \node at (0, 0) {\includegraphics[width=.7\columnwidth]{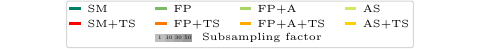}};
    \end{tikzpicture}

    \def\scaleimg{.36}
    \def\scaletxt{2*\scaleimg}
    
    \begin{subcaptiongroup}
        \subcaptionlistentry{}
        \label{fig:bf-c2sdl-hsc}
        \begin{tikzpicture}
            \node at (-.26\columnwidth, 0) {\includegraphics[width=\scaleimg\columnwidth]{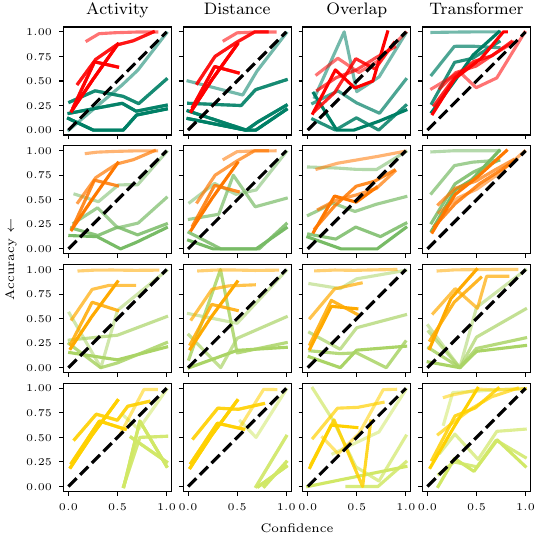}};
            \node at (-.65*\scaletxt*\columnwidth, 4.2*\scaletxt) {\captiontext*{}};
        \end{tikzpicture}~
        \subcaptionlistentry{}~
        \label{fig:bf-c2sdl-musc}~
        \hspace{.1\columnsep}~
        \begin{tikzpicture}
            \node at (-.26\columnwidth, 0) {\includegraphics[width=\scaleimg\columnwidth]{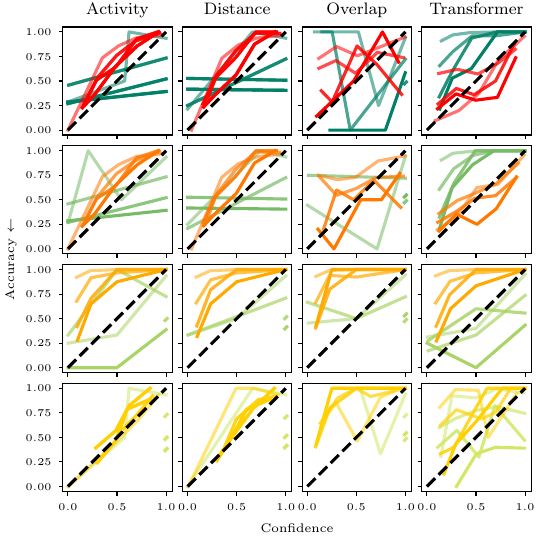}};
            \node at (-.65*\scaletxt*\columnwidth, 4.2*\scaletxt) {\captiontext*{}};
        \end{tikzpicture}
    \end{subcaptiongroup}
    \begin{subcaptiongroup}
        \subcaptionlistentry{}
        \label{fig:dic-c2dh-hela}
        \begin{tikzpicture}
            \node at (-.26\columnwidth, 0) {\includegraphics[width=\scaleimg\columnwidth]{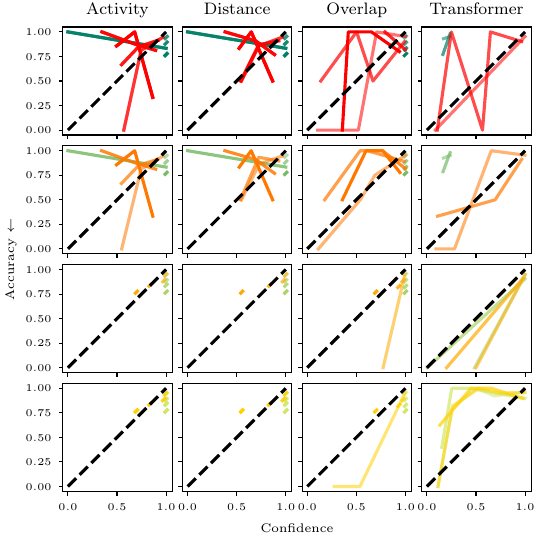}};
            \node at (-.65*\scaletxt*\columnwidth, 4.2*\scaletxt) {\captiontext*{}};
        \end{tikzpicture}~
        \subcaptionlistentry{}~
        \label{fig:fluo-n2dl-hela}~
        \hspace{.1\columnsep}~
        \begin{tikzpicture}
            \node at (-.26\columnwidth, 0) {\includegraphics[width=\scaleimg\columnwidth]{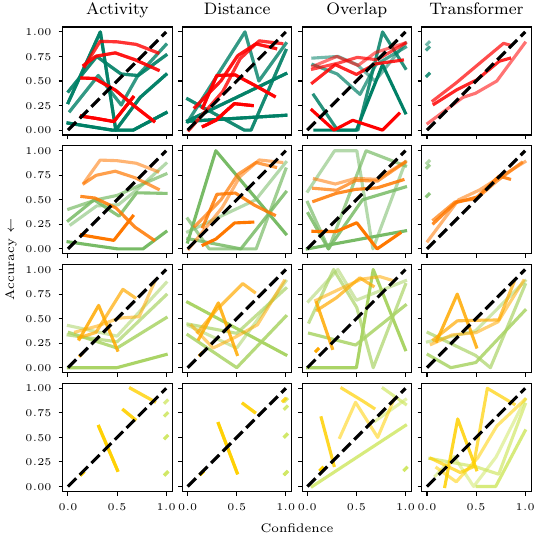}};
            \node at (-.65*\scaletxt*\columnwidth, 4.2*\scaletxt) {\captiontext*{}};
        \end{tikzpicture}
    \end{subcaptiongroup}
    \begin{subcaptiongroup}
        \subcaptionlistentry{}
        \label{fig:one-in-a-million}
        \begin{tikzpicture}
            \node at (-.26\columnwidth, 0) {\includegraphics[width=\scaleimg\columnwidth]{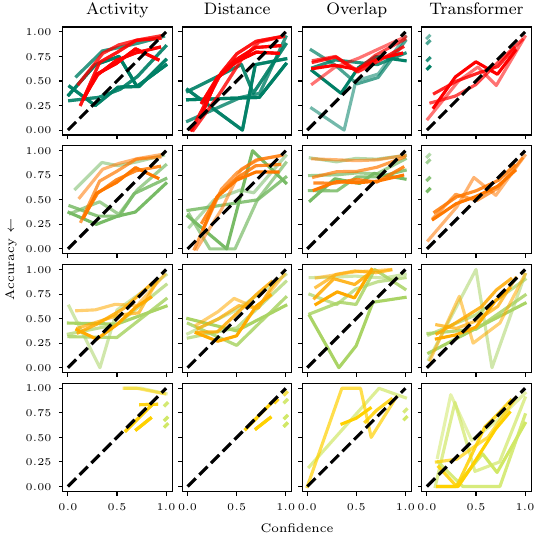}};
            \node at (-.65*\scaletxt*\columnwidth, 4.2*\scaletxt) {\captiontext*{}};
        \end{tikzpicture}~
        \subcaptionlistentry{}~
        \label{fig:phc-c2dl-psc}~
        \hspace{.1\columnsep}~
        \begin{tikzpicture}
            \node at (-.26\columnwidth, 0) {\includegraphics[width=\scaleimg\columnwidth]{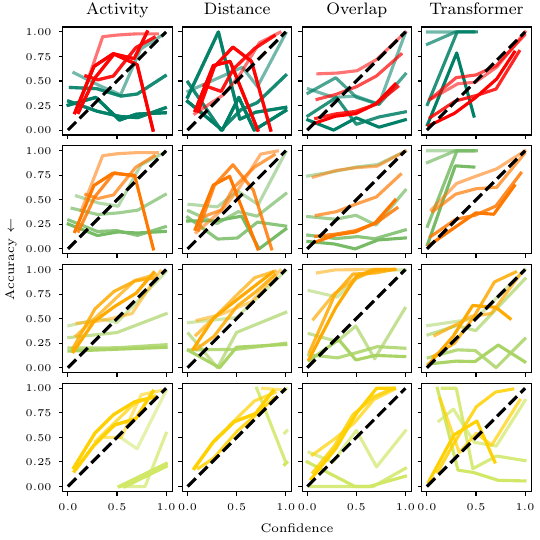}};
            \node at (-.65*\scaletxt*\columnwidth, 4.2*\scaletxt) {\captiontext*{}};

        \end{tikzpicture}
    \end{subcaptiongroup}
    \caption{
        Calibration curves of all tested methods and datasets. Datasets are 
        (a) BF-C2DL-HSC,
        (b) BF-C2DL-MuSC,
        (c) DIC-C2DH-HeLa,
        (d) Fluo-N2DL-HeLa,
        (e) PhC-C2DL-PSC all taken from the Cell Tracking Challenge \cite{maska_benchmark_2014, maska_cell_2023} and,
        (f) Tracking-One-in-a-Million dataset \cite{seiffarth_tracking_2024}.
        The gray dashed line depicts 1-to-1 correspondence, \ie perfect calibration.
    }
    \label{fig:caliration-detailed}
\end{figure*}
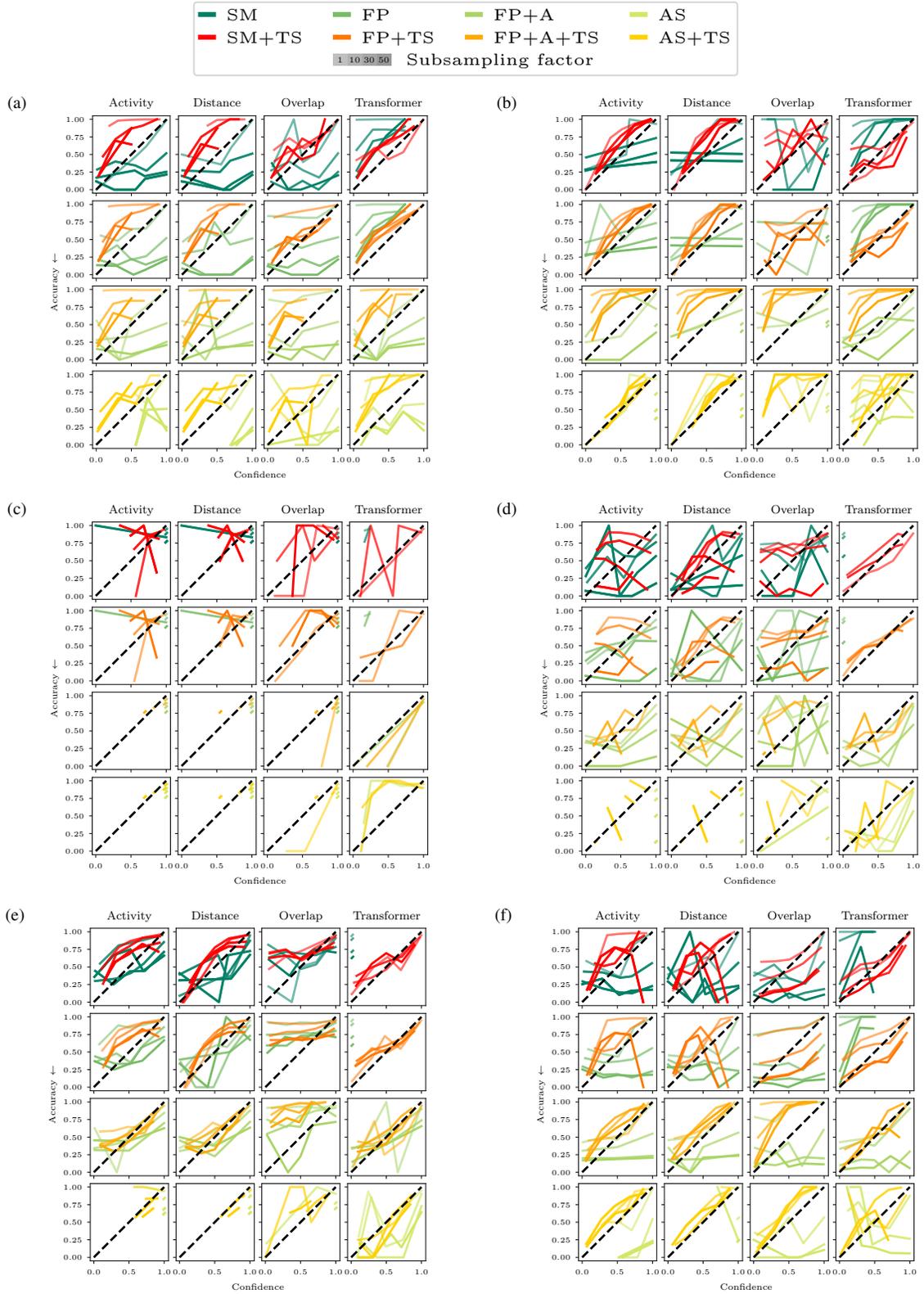

\end{document}